\documentclass[preprint,authoryear]{elsarticle}

\usepackage{geometry}
\usepackage{graphicx}
\usepackage{amsmath, bm}
\usepackage{booktabs}
\usepackage{algorithm}
\usepackage{algorithmic}
\usepackage{amsthm}
\usepackage{amssymb,bbm}

\usepackage{epstopdf}
\usepackage{tikz}
\usetikzlibrary{bayesnet}
\usepackage{algorithm}
\usepackage{algorithmic}
\usepackage{graphicx}
\usepackage{array}
\usepackage{hyperref}

\usepackage{caption}
\usepackage[normalem]{ulem}
\usepackage{soul}
\usepackage{subfig}
\usepackage{dsfont}
\usepackage{xcolor}

\usepackage{times}
\usepackage{blindtext}
\usepackage{latexsym}
\usepackage{amsmath}
\usepackage{amssymb}
\usepackage{dsfont}
\usepackage{graphicx}
\usepackage{subcaption}
\usepackage{threeparttable}
\usepackage[T1]{fontenc}
\usepackage[utf8]{inputenc}
\usepackage{microtype}
\usepackage{inconsolata}
\usepackage{stmaryrd}
\usepackage{multirow}
\usepackage{pdfpages}
\usepackage{newfloat}
\usepackage{listings}

\usepackage{booktabs}
\usepackage{cleveref}

\usepackage{amsthm}

\definecolor{lightblue}{rgb}{0.8, 0.9, 1.0}

\journal{Neurocomputing}

\begin{document}

\begin{frontmatter}



\title{{WAVE++: Capturing Within-Task Variance for Continual Relation Extraction with Adaptive Prompting}\tnoteref{t1}}
\tnotetext[t1]{A part of this work appears in \citet{le2024adaptive}.}



\author{Bao-Ngoc Dao\corref{equal}}
\ead{ngoc.db224884@sis.hust.edu.vn}

\author{Minh Le\corref{equal}}
\ead{minh611002@gmail.com}

\author{Quang Nguyen\corref{equal}}
\ead{quang.nm144@gmail.com}

\author{Luyen Ngo Dinh\corref{equal}}
\ead{ngodinhluyennht@gmail.com}

\author{Nam Le\corref{corr}}
\ead{namlh@soict.hust.edu.vn}

\author{Linh Ngo Van}
\ead{linhnv@soict.hust.edu.vn}

\address{Hanoi University of Science and Technology, No. 1, Dai Co Viet road, Hanoi, Vietnam}



\cortext[equal]{Equal contribution}
\cortext[corr]{Corresponding author}



\begin{abstract}
Memory-based approaches have shown strong performance in Continual Relation Extraction (CRE). However, storing examples from previous tasks increases memory usage and raises privacy concerns. Recently, prompt-based methods have emerged as a promising alternative, as they do not rely on storing past samples. Despite this progress, current prompt-based techniques face several core challenges in CRE, particularly in accurately identifying task identities and mitigating catastrophic forgetting. Existing prompt selection strategies often suffer from inaccuracies, lack robust mechanisms to prevent forgetting in shared parameters, and struggle to handle both cross-task and within-task variations. 
In this paper, we propose \textbf{WAVE++}, a novel approach inspired by the connection between prefix-tuning and mixture of experts. Specifically, we introduce task-specific prompt pools that enhance flexibility and adaptability across diverse tasks while avoiding boundary-spanning risks; this design more effectively captures both within-task and cross-task variations. To further refine relation classification, we incorporate label descriptions that provide richer, more global context, enabling the model to better distinguish among different relations. We also propose a training-free mechanism to improve task prediction during inference. Moreover, we integrate a generative model to consolidate prior knowledge within the shared parameters, thereby removing the need for explicit data storage. Extensive experiments demonstrate that WAVE++ outperforms state-of-the-art prompt-based and rehearsal-based methods, offering a more robust solution for continual relation extraction.
Our code is publicly available at \url{https://github.com/PiDinosauR2804/WAVE-CRE-PLUS-PLUS}.
\end{abstract}


\end{frontmatter}



\theoremstyle{plain}
\theoremstyle{definition}
\theoremstyle{remark}

\theoremstyle{plain}
\newtheorem{theorem}{Theorem}[section]
\newtheorem{proposition}[theorem]{Proposition}
\newtheorem{lemma}[theorem]{Lemma}
\newtheorem{corollary}[theorem]{Corollary}
\theoremstyle{definition}
\newtheorem{definition}[theorem]{Definition}
\newtheorem{assumption}[theorem]{Assumption}
\theoremstyle{remark}
\newtheorem{remark}[theorem]{Remark}

\def\ie{{\em i.e.,~}}
\def\eg{{\em e.g.,~}}

\newcommand{\dbm}{{\bm d}}

\newcommand{\xbm}{{\bm x}}
\newcommand{\Xbm}{{\bm X}}

\newcommand{\kbm}{{\bm k}}
\newcommand{\hbm}{{\bm h}}
\newcommand{\htil}{{\Tilde{\hbm}}}

\newcommand{\Pbm}{{\bm P}}
\newcommand{\pbm}{{\bm p}}

\newcommand{\Xbf}{{\mathbf{X}}}
\newcommand{\Pbf}{{\mathbf{P}}}

\newcommand{\ybf}{{\mathbf{y}}}

\newcommand{\Pbb}{{\mathbb{P}}}

\def\RR{\mathbb{R}}

\newcommand{\softmax}{\mathrm{softmax}}

\newcommand{\zbm}{{\bm z}}

\newcommand{\cov}{\boldsymbol{\Sigma}}
\newcommand{\mean}{\boldsymbol{\mu}}

\newcommand{\dv}{d_v}

\section{Introduction} \label{sec:introduction}

\emph{Continual Relation Extraction (CRE)} involves training models to progressively extract relationships between entity pairs across a sequence of tasks \citep{wang2019sentence, zhao2022consistent, nguyen2023spectral, le2024adaptive}. As a specialized domain of continual learning, CRE's primary goal is to mitigate \emph{catastrophic forgetting} \citep{mccloskey1989catastrophic, nguyen2019toward}, where model performance deteriorates as the number of tasks increases. To address this issue, most CRE methods \emph{employ a memory buffer}, which {stores samples to revisit previous tasks and preserve learned knowledge} \citep{han2020continual, zhao2022consistent}. Approaches utilizing this strategy are commonly referred to as rehearsal-based methods.

Although rehearsal-based methods have shown notable success in mitigating catastrophic forgetting, they still exhibit significant limitations that demand more robust solutions. First, despite the use of memory buffers, the representations of learned relations often degrade shortly after training shifts to subsequent tasks \citep{caccia2021new}. Second, these methods violate a core principle of continual learning, {by necessitating ongoing access to data from previous tasks}. This reliance raises substantial concerns about \emph{data privacy} and the considerable \emph{storage demands} of large-scale memory buffers. {A pressing need therefore exists for} alternative approaches that minimize memory usage, thereby preserving the principles of continual learning while {addressing critical privacy and scalability challenges} \citep{ke2022continual}.

Recent advances in continual learning take inspiration from prompt-based techniques in natural language processing, leading to a new class of methods. These methods leverage learnable parameters, known as \emph{prompts}, to guide pre-trained models on downstream tasks without requiring access to past data \citep{wang2022learning, wang2022dualprompt, wang2024hierarchical}. Unlike traditional memory-replay methods, these approaches \emph{do not rely on storing samples from previous tasks}; instead, they insert small sets of auxiliary parameters to steer the training process. These prompts are adaptable to specific tasks, enabling continual relation extraction without the need for data replay. 

Despite these advantages, current prompt-based frameworks still exhibit several limitations. First, they do not fully alleviate catastrophic forgetting because many approaches \emph{rely on shared components}, such as the Prompt Pool \citep{wang2022learning}, the General Prompt (G-Prompt) \citep{wang2022dualprompt}, or a shared MLP classifier, thereby constraining their adaptability to varying task distributions. Second, the task-prediction mechanisms employed by these methods \citep{wang2022dualprompt, wang2024hierarchical} are \emph{prone to misclassification errors}, as any mismatch between the prompts used during training and those employed at inference can degrade performance. {Finally, prompt-based methods often \emph{fail to address both cross-task and within-task variability}. For example, \citet{wang2022learning} introduce a single, shared prompt pool, which can result in inputs from different relations using the same or very similar prompts. This limitation is especially prominent in continual relation extraction, where instances from different relation classes may present almost identical contexts, as illustrated by the following examples:
\begin{itemize}
    \item \texttt{"[X] is a student at [Z college]"}. 
    \item \texttt{"[X] supervises a student at [Z college]"}. 
\end{itemize}
When such instances share the same prompts, their final hidden representations may collapse—becoming too similar for the relation classifier head to distinguish between them effectively.}

In alignment with these approaches, \citet{le2024mixture} investigate the connection between \emph{Prefix-tuning} \citep{li2021prefix}, a widely adopted prompt-based technique, and \emph{Mixture of Experts (MoE)} models \citep{Jacob_Jordan-1991, jordan1994hierarchical}. Their findings reveal that \emph{self-attention can be interpreted as containing multiple MoE models}, and that implementing prefix-tuning is equivalent to adding new \emph{prefix} experts to these pre-trained MoE architectures, thereby enabling the fine-tuning of underlying representations.

Building on this insight, we propose \textbf{WAVE++} (\underline{W}ithin-T\underline{a}sk \underline{V}ariance \underline{A}war\underline{e}ness for CRE), a prompt-based method designed to address the prior limitations. Rather than employing a single prompt pool across all tasks, WAVE++ assigns \emph{a dedicated prompt pool for each task}, thereby improving plasticity in response to task distribution shifts and capturing task-specific characteristics. Additionally, WAVE++ integrates \emph{label descriptions} of relations to learn global, relation-specific contexts, ensuring that essential characteristics of each relation are preserved.

To mitigate catastrophic forgetting in the shared parameters, WAVE++ incorporates \emph{generative models} that produce instructed latent data representations for replay. This strategy guards against representation drift without requiring the storage of extensive raw data, offering a more efficient and comprehensive alternative to conventional replay-based methods. Furthermore, although recent approaches often rely on a classifier head for task prediction \citep{wang2024hierarchical, le2024adaptive}, WAVE++ removes this dependency by \emph{selecting the relevant prompt pool through a simple voting mechanism}. Experimental results show that WAVE++ achieves state-of-the-art performance, outperforming existing prompt-based and rehearsal-based baselines.

\paragraph{Contributions} Our main contributions can be summarized as follows: 
\begin{itemize} 
\item We identify key limitations in current prompt-based continual learning approaches, including ineffective prompt selection, inadequate mitigation of catastrophic forgetting in shared parameters, and suboptimal strategies for handling both cross-task and within-task variance. 

\item  We highlight the underlying connections between mixture of experts models and prompt-based continual learning techniques. Specifically, we show that any prompt-based continual learning framework can be viewed as a specialized instance of MoE architectures. 

\item We propose \textbf{WAVE++}, which addresses the identified shortcomings by introducing task-specific prompt pools, leveraging relation label descriptions, employing a voting-based task prediction mechanism, and incorporating generative modeling of latent representations to enhance CRE performance. 

\item {We conduct extensive experiments demonstrating that} WAVE++ outperforms state-of-the-art prompt-based and rehearsal-based methods, validating both its effectiveness and versatility. 
\end{itemize}



\begin{figure}[t]
    \centering
    \includegraphics[width=0.55\columnwidth]{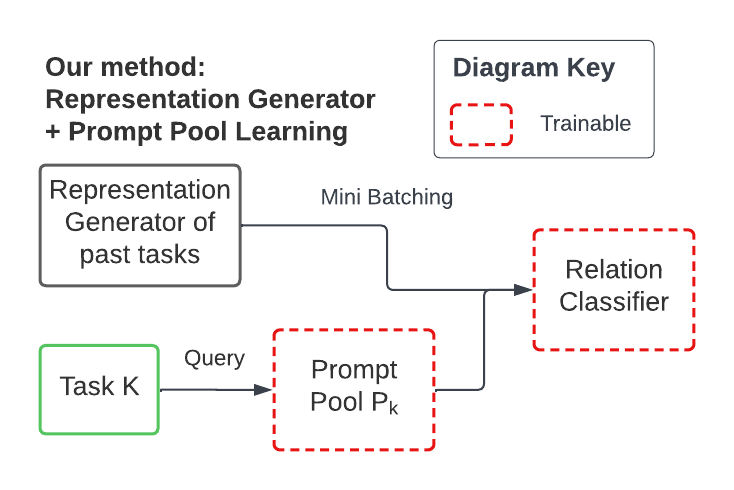} 
    \caption{Overall framework  of WAVE++.
    To mitigate forgetting across tasks, each $k$-th task is associated with its own prompt pool $\mathcal{P}_k$, rather than relying on a single, shared prompt pool as in L2P. In addition, we employ representation generators to synthesize information from previously learned tasks, thereby reinforcing {the capacity of the relation classifier} to retain accumulated knowledge.
    }
    \label{fig:Our framework}
    \vskip -0.1in
\end{figure}

\section{Background and Related Work} \label{sec:background}

In this section, we briefly {review the relevant background of} continual relation extraction in Section~\ref{sec:background_cre}. We then examine recent advancements in prompt-based approaches, followed by an overview of MoE models in Section~\ref{sec:background_prompt} and Section~\ref{sec:background_moe}, respectively.

\subsection{Continual Relation Extraction} \label{sec:background_cre}

\begin{figure}[t]
\vskip -0.3in
\begin{center}
\centerline{\includegraphics[width=\columnwidth]{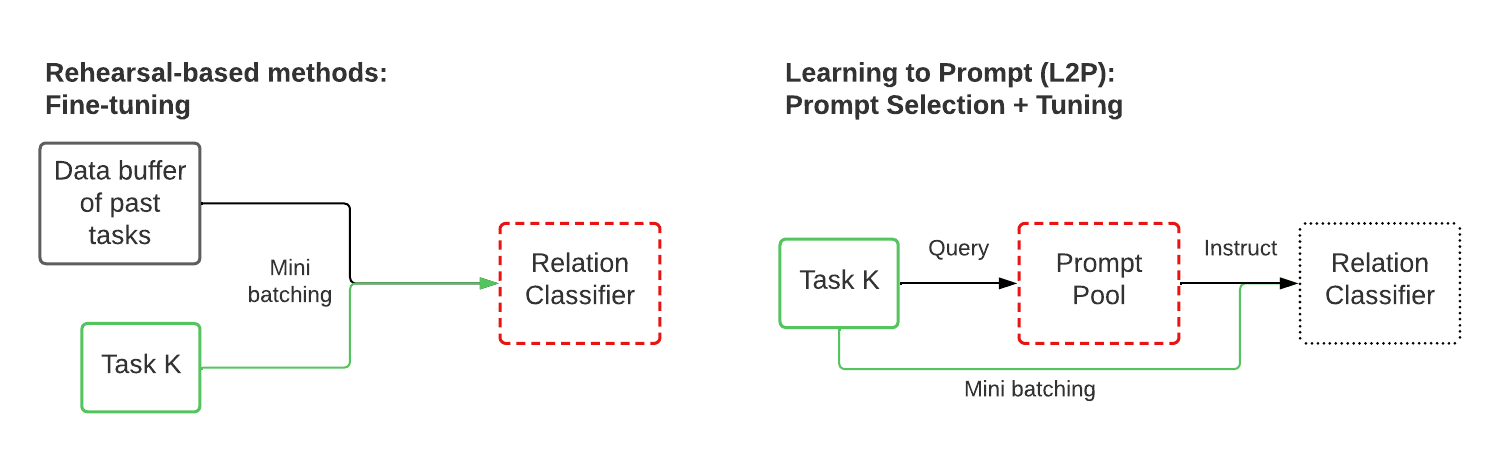}}
\caption{Comparison with rehearsal-based methods. Unlike approaches that rely on a rehearsal buffer, L2P employs a single backbone model and a prompt pool to store task-specific knowledge, thereby eliminating the need for explicit rehearsal to prevent catastrophic forgetting. L2P further adapts to each instance by selecting and updating prompts from the pool on a per-instance basis.}
\label{fig:l2p}
\end{center}
\vskip -0.4in
\end{figure}

{\emph{Continual relation extraction} (CRE) is a {special subfield} of relation extraction that addresses the challenge of incrementally learning new relational patterns from incoming tasks while preserving knowledge of previously learned relations.} Formally, CRE involves learning sequentially from a series of tasks $\{\mathcal{T}_1,..., \mathcal{T}_k\}$, where $\mathcal{T}_t$ denotes the $t$-th task. Similar to traditional supervised classification frameworks~\citep{ji2020span}, each task $\mathcal{T}_t$ is associated with a dataset $\mathcal{D}_t=\{(\xbm^t_i, y^t_i) \ | \ y^t_i \in \mathcal{R}_t \}_{i=1}^{\mathcal{N}_t}$ containing $\mathcal{N}_t$ input–label pairs and a distinct set of relations $\mathcal{R}_t$. Notably, these relation sets are mutually exclusive across tasks, such that $\mathcal{R}_i \bigcap \mathcal{R}_j=\emptyset$ for all $i, j \in \{1,\dots,k\}$ where $i \ne j$. Furthermore, once the model completes training on a given task $t$, it can no longer access the data from that task when learning subsequent tasks.

As training proceeds, the model must maintain its performance on all relations encountered so far, represented as $\widehat{\mathcal{R}}_t=\bigcup_{i=1}^t \mathcal{R}_i$. A primary challenge under these constraints is \emph{catastrophic forgetting}, wherein performance on earlier tasks deteriorates as the model adapts to new ones \citep{mccloskey1989catastrophic, nguyen2019toward}. Many strategies have been explored to address this challenge. For instance, EA-EMR \citep{wang2019sentence} combines memory replay and embedding alignment to alleviate catastrophic forgetting. RP-CRE \citep{cui2021refining} employs a memory network to refine sample embeddings with relation prototypes. Similarly, CRE-DAS \citep{zhao2023improving} introduces memory-insensitive relation prototypes and memory augmentation to mitigate overfitting. CDec+ACA \citep{xia2023enhancing} proposes a classifier decomposition framework to address representation biases by learning robust representations while preserving previously acquired knowledge. However, \emph{most existing approaches for CRE rely on a memory buffer to store samples of previously encountered relations}, which raises critical concerns regarding memory overhead and privacy (see Figure~\ref{fig:l2p}, left).

Recently, \citet{kim2022theoretical} presented a theoretical framework that decomposes the continual learning objective into more tractable components, offering a structured approach to mitigate catastrophic forgetting. Specifically, given a sample $(\xbm, y)$ at test time with $y \in \mathcal{R}_t$, the model predicts the relation of this sample as $\hat{y}$. The probability of predicting the correct label, $\Pbb\left(\hat{y} = y \ | \ \xbm\right)$, is then factorized into two components: \emph{within-task prediction} (WTP) and \emph{task-identity inference} (TII), as follows:
\begin{align}
    \Pbb\left(\hat{y} = y \ | \ \xbm\right) = 
    \underbrace{\Pbb\left(\hat{y} \in \mathcal{R}_t \ | \ \xbm\right)}_\text{Task-identity inference}
    \times
    \underbrace{\Pbb\left(\hat{y} = y \ | \ \hat{y} \in \mathcal{R}_t ,\xbm\right)}_\text{Within-task prediction}
    ,\label{eq:cil_decompostion}
\end{align}
where the first term refers to the model’s ability to determine the appropriate task, and the second term represents its ability to accurately predict the label within that task. It has been shown that improving these two components is both a \emph{necessary} and \emph{sufficient} condition for enhancing overall continual learning performance. This framework provides a solid foundation for designing more effective rehearsal-free continual learning algorithms~\citep{wang2024hierarchical, le2024mixture}. In this work, we build upon this theoretical framework by proposing methods to improve both TII and WTP (see Section~\ref{sec:method} for details).

\subsection{Prompt-based Approaches}
\label{sec:background_prompt}

Owing to robust generalization capabilities, recent studies have increasingly leveraged pre-trained models to mitigate catastrophic forgetting, achieving impressive performance gains~\citep{wang2022learning, wang2023isolation, wang2023rehearsal, zhou2024ensemble}. {Following previous research~\citep{xia2023enhancing, zhao2023improving}, we adopt a pre-trained BERT model~\citep{devlin2018bert} as our base architecture. However, the conventional approach of fine-tuning all model parameters presents substantial computational challenges.} To address this challenge, \emph{Parameter Efficient Fine-Tuning} (PEFT) strategies have been proposed as an alternative paradigm~\citep{houlsby2019parameter, hu2021lora, lester2021power, li2021prefix, le2025adaptive}. PEFT focuses on designing fine-tuning methods that update only a small subset of parameters while leaving the rest of the model unchanged. By doing so, it significantly reduces the computational overhead associated with full fine-tuning. Among these techniques, \emph{prompt-tuning}~\citep{lester2021power} and \emph{prefix-tuning}~\citep{li2021prefix} have emerged as simple yet effective methods. Both approaches append learnable \emph{prompt} tokens to the input, either at the input layer alone or within intermediate representations, to serve as instructions for the pre-trained Transformer model to adapt to downstream tasks. In this work, we adopt prefix-tuning as our prompting approach. Specifically, the Transformer architecture comprises multiple consecutive \emph{multi-head self-attention} (MSA) layers~\citep{vaswani2017attention}, which are defined as follows:
\begin{definition}[Multi-head Self-Attention Layer]
    Let $\Xbf^Q = \Xbf^K = \Xbf^V = \left[ \xbm_1, \dots, \xbm_N \right]^\top \in \RR^{N \times d}$ be the input query, key, and value matrices, where $N$ is the sequence length and $d$ is the embedding dimension. The MSA layer output is given by:
    \begin{align}
        &\mathrm{MSA}(\Xbf^Q, \Xbf^K , \Xbf^V) = \mathrm{Concat}(\hbm_1, ..., \hbm_{m})W^O \in \RR^{N \times d}, \nonumber \\
        &\hbm_i = \mathrm{Attention}(\Xbf^Q W_i^Q, \Xbf^K W_i^K, \Xbf^V W_i^V) \in \RR^{N \times d_v}, \label{eq:msa}
    \end{align}
    where $m$ is the number of attention heads, and $W^Q_i \in \RR^{d \times d_k}$, $W^K_i \in \RR^{d \times d_k}$, $W^V_i \in \RR^{d \times d_v}$, $W^O \in \RR^{md_v \times d}$ are the projection matrices. Typically $d_k = d_v = d / m$ so that the total output dimension remains $d$.
\end{definition}
Building on the attention mechanism, prefix-tuning introduces learnable prompt parameters at the inputs of these MSA layers, enabling efficient fine-tuning of the pre-trained Transformer model.
\begin{definition}[Prefix-tuning]
    Let $\Pbf = \left[\Pbf^K; \Pbf^V \right]$, where $\Pbf^K = \left[ \pbm_1^K, \dots, \pbm_L^K \right]^\top \in \RR^{L \times d}$ and $\Pbf^V = \left[ \pbm_1^V, \dots, \pbm_L^V \right]^\top \in \RR^{L \times d}$ are the prefix key and prefix value parameters, respectively. Prefix-tuning injects these parameters into the key and value matrices:
    \begin{align}
        f^{\mathrm{Pre-T}}_{\mathrm{prompt}}(\Pbf, \Xbf^Q, \Xbf^K , \Xbf^V) &= \mathrm{MSA}\left(
        \Xbf^Q, 
        \begin{bmatrix}
            \Pbf^K \\ \Xbf^K
        \end{bmatrix},
        \begin{bmatrix}
            \Pbf^V \\ \Xbf^V
        \end{bmatrix}
        \right) 
        \in \RR^{N \times d}. \label{eq:prefix-tuning}
    \end{align}
\end{definition}
{By not prepending prompts to the query matrix, prefix-tuning preserves the original input sequence dimension.} Importantly, only the prompt parameters $\Pbf$ are trained, while the parameters of the pre-trained Transformer backbone including $W^Q_i$, $W^K_i$, $W^V_i$, and $W^O$ remain frozen. {Inspired by these prompting techniques, several continual learning methods, such as L2P~\citep{wang2022learning}, DualPrompt~\citep{wang2022dualprompt}, and HiDe-Prompt~\citep{wang2024hierarchical}, mitigate catastrophic forgetting without requiring historical data. These methods are discussed in greater detail in Section~\ref{sec:viewpoint}.}

\subsection{Mixture of Experts}\label{sec:background_moe}

\emph{Mixture of Experts} (MoE) is a classical ensemble learning framework that combines multiple models to achieve more expressive and accurate predictions~\citep{Jacob_Jordan-1991, jordan1994hierarchical}.  Formally, an MoE model consists of $N$ \emph{expert networks}, denoted by $f_i: \mathbb{R}^d \rightarrow \mathbb{R}^{d_v}$ for $i = 1, \dots, N$. A \emph{gating function}, $G: \mathbb{R}^d \rightarrow \mathbb{R}^{N}$, dynamically determines the contribution of each expert for a given input $\xbm \in \RR^d$. The gating function is defined by a set of learned \emph{score functions}, $s_i: \mathbb{R}^d \rightarrow \mathbb{R}$, associated with each expert. {The final output is formulated as:}
\begin{align*}
    \ybf &= \sum_{j=1}^N G(\xbm)_j \cdot f_j(\xbm) = \sum_{j=1}^N \frac{\exp\left(s_j(\xbm)\right)}{\sum_{\ell=1}^N\exp\left(s_\ell(\xbm)\right)} \cdot f_j(\xbm).
\end{align*}
Building on this concept, \citet{shazeer2017outrageously} proposed the \emph{Sparse Mixture of Experts} (SMoE) architecture to efficiently scale large models. This is achieved by {using a sparse gating function $\mathrm{TopK}$, which selects the $K$ experts with the highest scores $s_j(\xbm)$ and assigns a score of $-\infty$ to the remaining $N-K$ experts}. Formally, the $\mathrm{TopK}$ function is defined as:
\begin{align*}
\mathrm{TopK}\left(\bm v, K\right)_i
 =\left\{\begin{array}{cc}
\bm v_i, & \text { if } \bm v_i \text { is in the } K \text { largest elements of } \bm v \\
-\infty, & \text{ otherwise. }
\end{array}\right.
\end{align*}
After the gating function selects the top $K$ experts, their outputs are linearly combined using their normalized scores:
\begin{align*}
\ybf=\sum_{j=1}^N \softmax\left(\mathrm{TopK}\left(s(\xbm), K\right)\right)_j \cdot  f_j(\xbm),
\end{align*}
where $s(\xbm) = (s_1(\xbm),\dots, s_N(\xbm))$. MoE models have attracted considerable attention due to their flexibility and adaptability across various domains, including large language models~\citep{du2022glam, zhou2023brainformers,li2024cumo}, computer vision~\citep{riquelme2021scaling}, and multi-task learning~\citep{ma2018modeling}.



\begin{figure}[t]
\begin{center}
\centerline{\includegraphics[width=0.9\columnwidth]{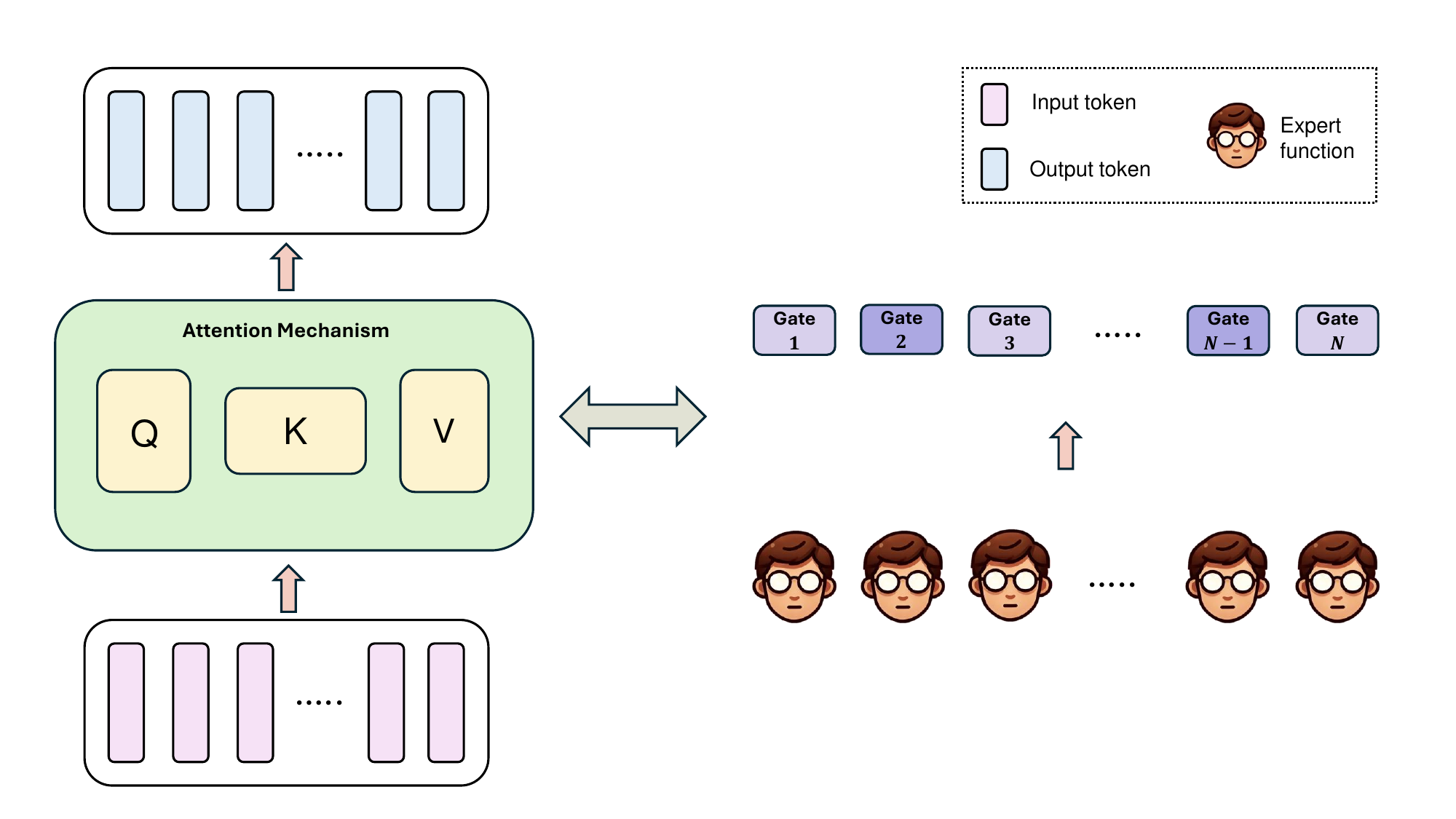}}
\caption{Illustration of the connection between attention and mixture of experts. An attention head can be interpreted as a collection of MoE modules that share a common set of experts but employ distinct gating functions. This structure closely parallels the multi-gate MoE architecture.}
\label{fig:attention_moe}
\end{center}
\vskip -0.4in
\end{figure}

\section{From Mixture of Experts to Prompt-based Continual Learning}
\label{sec:viewpoint}

In this section, we discuss the relationship between mixture of experts and prompt-based techniques, demonstrating how existing prompt-based continual learning methods can be interpreted within this framework. Specifically, recent work~\citep{le2024mixture} has shown that each output vector from each attention head in an MSA layer can be viewed as the output of an MoE model. This perspective implies that an MSA layer can be regarded as a special architecture in which \emph{each attention head comprises multiple MoE models}. Furthermore, these studies suggest that prefix-tuning can be {interpreted as a method of fine-tuning these models by \emph{introducing new experts}}.

\begin{proposition} \label{props: attn_moe}
Let $\Xbm = \left[\xbm_1^\top,\dots,\xbm_N^\top\right]^\top \in \RR^{Nd}$ denote the concatenation of all input token embeddings. From Equation~\eqref{eq:msa}, the output of the $l$-th attention head is $\hbm_l = \left[ \hbm_{l, 1}, \dots, \hbm_{l, N} \right]^\top
\in \RR^{N \times d_v}$. Each output vector $\hbm_{l, i} \in \RR^{d_v}$ can then be expressed as an MoE model:
\begin{align}
    \hbm_{l,i} &= \sum^N_{j=1} \frac{\exp(s_{i,j}(\Xbm))}{\sum^N_{k=1} \exp({s_{i,k}}(\Xbm))}\cdot f_j(\Xbm),
\end{align}
where the expert function is $f_j: \RR^{Nd} \rightarrow \RR^{d_v}$, and the score function is $s_{i, j}: \RR^{Nd} \rightarrow \RR$. Both functions take the full input sequence $\Xbm$ as input. For more details, please refer to~\ref{appendix:proof_of_attn_moe}.
\end{proposition}

This structure is analogous to the \emph{Multi-gate Mixture of Experts} framework~\citep{ma2018modeling}, where multiple MoE models share a common set of experts but use distinct gating functions, as illustrated in Figure~\ref{fig:attention_moe}. {In a pre-trained model, all the parameters of the experts $f_j$ and score functions $s_{i, j}$ are frozen. Consequently, these pre-trained experts $f_1, \dots, f_N$ can be considered a knowledge base.} Prefix-tuning augments this knowledge base by introducing additional \emph{prefix experts} encoded as prompt tokens.

{
\begin{proposition} \label{props: prompt_moe}
    From Equation~\eqref{eq:prefix-tuning}, the output of the $l$-th attention head modified by prefix-tuning can be written as:
\begin{align}
    \htil_l = \mathrm{Attention}\left(
    \Xbf^Q W_l^Q, 
    \begin{bmatrix}
            \Pbf^K \\ \Xbf^K
    \end{bmatrix} W_l^K, 
    \begin{bmatrix}
            \Pbf^V \\ \Xbf^V
    \end{bmatrix} W_l^V
    \right) = \left[ \htil_{l, 1}, \dots, \htil_{l, N} \right]^\top
    \in \RR^{N \times d_v}, \label{eq:prefix_tunining_output}
\end{align}
where each output vector $\htil_{l, i} \in \RR^{d_v}$ can be expressed as an MoE model:
\begin{align}
    \htil_{l,i} 
    &= \sum_{j = 1}^N  
        \underbrace{\frac{\exp(s_{i, j}(\Xbm))}
        {
            \sum_{k = 1}^{N + L} \exp(s_{i, k}(\Xbm))
        } f_j(\Xbm)}_\text{pre-trained experts}
    + \sum_{j' = 1}^L  
        \underbrace{\frac{\exp(s_{i, N + j'}(\Xbm))}
        {
            \sum_{k = 1}^{N + L} \exp(s_{i, k}(\Xbm))
        } f_{N + j'}(\Xbm)}_\text{new prefix experts}
    , \label{eq:prefix_moe}
\end{align}
for $i = 1, \dots, N$. Here, the new expert functions $f_{N + j'}: \RR^{Nd} \rightarrow \RR^{d_v}$, and score functions $s_{i, N + j'}: \RR^{Nd} \rightarrow \RR$ are functions of $\Xbm$ with parameters derived from $\Pbf^K$ and $\Pbf^V$. For more details, please refer to~\ref{appendix:proof_of_prompt_moe}.
\end{proposition}
}

Hence, prefix-tuning can be interpreted as adding new experts $f_{N + 1}, \dots, f_{N + L}$ to the original pre-trained experts $f_1, \dots, f_N$. These additional experts collaborate with the original ones to adapt the model for downstream tasks, thus alleviating the need to retrain all of its parameters. This perspective offers {a new avenue for analyzing and exploring existing prompt-based continual learning methods.}

Here, we take L2P~\citep{wang2022learning} as a concrete example. L2P introduces the concept of a \emph{Prompt Pool}, consisting of multiple prompts that guide the pre-trained model to adapt to all tasks. Specifically, let the pool of $M$ prompts be
\begin{align}
    \mathcal{P} = \{(\kbm_1, \Pbm_1),\dots, (\kbm_M, \Pbm_M) \}, \label{eq:l2p_prompt_pool}
\end{align}
where each prompt $\Pbm_i \in \RR^{L \times d}$ is associated with a learnable key $\kbm_i \in \RR^d$ for $i = 1, \dots, M$. Given an input instance $\xbm$, we first compute a query vector $q(\xbm) \in \RR^d$ using the pre-trained transformer model. We then employ a query-key matching function $\gamma: \mathbb{R}^d \times \mathbb{R}^d \rightarrow \mathbb{R}$ (\eg cosine similarity) to select top-$K$ most relevant prompts by solving the following objective:
\begin{align}  
\boldsymbol{K}_{\boldsymbol{x}} = \underset{S \subseteq \{1, \dots, M\}: |S| = K}{\rm argmin} \sum_{s \in S}\gamma(q(\boldsymbol{x}), \boldsymbol{k}_{s}), \label{eq:L2P}
\end{align}
where $\boldsymbol{K}_{\boldsymbol{x}}$ is the set of indices for the top-$K$ keys most relevant to $\boldsymbol{x}$. The corresponding prompts are then prepended to $\xbm$ to form the prompted input $\xbm_p$, which is finally fed into the transformer model to generate the final prediction. By serving as a repository of task-specific knowledge, the prompt pool removes the need for a rehearsal buffer to prevent catastrophic forgetting (see Figure~\ref{fig:l2p}, right). 

The prompt pool can be viewed as a collection of $M \cdot L$ experts $f_{N + 1}, \dots, f_{N + ML}$. Each prompt $\Pbm_i$ implicitly encodes a group of $L$ experts, $f_{N + (i-1)L + 1}, \dots, f_{N + iL}$ for $i = 1, \dots, M$. Consequently, selecting $K$ prompts from the pool is analogous to selecting $K \times L$ experts. Each expert is assigned based on the query feature $q(\xbm)$, which captures contextual information. However, because each prompt $\Pbm_i$ is associated with a single prompt key $\kbm_i$, \emph{all $L$ experts within a given prompt are tied to a single prompt key $\kbm_i$}, potentially limiting the model’s expressiveness. Moreover, employing a single prompt pool for all tasks can lead to issues such as forgetting or noisy knowledge, as \emph{instances from different tasks may share certain experts}, thereby reducing cross-task variability.

DualPrompt \citep{wang2022dualprompt} enhances L2P by employing two complementary prompts: a general prompt (G-Prompt) and a task-specific expert prompt (E-Prompt) for each task. {{When a new task arrives, only the G-Prompt and the newly introduced E-Prompt for that task are updated during training, while E-Prompts from previous tasks remain fixed. The set of E-Prompts serves as an expanding pool of task-specific knowledge, similar to the L2P prompt pool, with the key difference that it grows incrementally with each new task. DualPrompt retains L2P’s prompt selection mechanism for E-Prompts. In contrast, the G-Prompt is shared among all tasks and thus requires no selection. We posit that the G-Prompt aims to augment the set of pre-trained experts $f_1, \dots, f_N$ by introducing additional experts that capture generalizable knowledge. However, unlike the pre-trained experts, these newly learned experts in the G-Prompt are continually updated across tasks, which makes them susceptible to forgetting.}

HiDe-Prompt~\citep{wang2024hierarchical}, by comparison, uses only task-specific E-Prompts.  For prompt selection, it leverages an auxiliary MLP head trained with a cross-entropy loss to select the appropriate prompt, \emph{treating each task as a separate class}. This class-based approach may be suboptimal because the classes are defined by task order rather than semantic distinctions. Moreover, \emph{restricting each task to a single prompt may be insufficient to capture its full complexity}, given the limited expressiveness of prefix-tuning \citep{petrov2023prompting}. Specifically, prefix experts represent offset vectors encoded by prompt tokens, rather than the linear functions used by pre-trained experts, as indicated in Equation~\eqref{eq:pretrain_expert} and Equation~\eqref{eq:prompt_expert}. This simplicity limits their capacity to accommodate the full range of variations within each task. We address these concerns in Section~\ref{sec:method}.

\section{Proposed Method} \label{sec:method}

In this section, we present our proposed method WAVE++ in detail. First, in Section~\ref{sec:method_prompt_pool}, we examine the relationship between the mixture of experts and prefix-tuning to develop task-specific prompt pools that capture the inherent variations within individual tasks. Next, in Section~\ref{sec:method_label_desc}, we introduce label descriptions combined with a contrastive loss to enhance the robustness of each prompt pool by encapsulating the essential characteristics of task relations. These two components collectively improve within-task performance and contribute to enhanced continual learning, as demonstrated in Equation~\eqref{eq:cil_decompostion}. Lastly, we employ cascade voting for task prediction (Section~\ref{sec:method_gen_model}) and use generative models to {mitigate classification bias toward previously learned tasks in the relation classifier (Section~\ref{sec:method_tap}). Figure~\ref{fig:Our framework} provides an overview of our complete framework.}

\subsection{Task-specific Prompt Pool} \label{sec:method_prompt_pool}

As discussed in Section~\ref{sec:viewpoint}, recent advances in continual learning, such as HiDe-Prompt~\citep{wang2024hierarchical}, typically employ a dedicated prompt for each task. Although these task-specific prompts help mitigate catastrophic forgetting by leveraging specialized features, relying on a single prompt for all instances within a task can be overly restrictive. In particular, these prefix experts are simply constant functions, rendering them invariant to input variations. We posit that this limited expressiveness hinders the model from capturing the full complexity and variability inherent in each task. 

A straightforward strategy to enhance the expressiveness of prefix experts is to replace the constant function with a more complex one, such as a linear function. However, a key advantage of the original design {is its} simplicity and cost-effectiveness. {Naively implementing a more expressive prefix could significantly increase computational overhead, thereby undermining this core benefit.} An alternative approach, exemplified by L2P~\citep{wang2022learning}, involves maintaining a prompt pool from which prompts are selected based on the query features $q(\xbm)$. {This method adaptively chooses a set of experts based on the input, yielding a configuration that resembles a SMoE architecture. This design achieves greater expressiveness while retaining computational efficiency.}

Despite these advantages, repeatedly using the same prompts across different tasks increases the risk of interference. Such overlap can overwrite task-specific knowledge, erasing important information. To address this, we propose constructing \emph{a dedicated prompt pool for each task}. {This design isolates prompts on a per-task basis, thereby mitigating undesired overlap.} Specifically, for the $t$-th task, following the formulation presented in Equation~\eqref{eq:l2p_prompt_pool}, we introduce a prompt pool $\mathcal{P}_t$ consisting of $M$ prompts, defined as follows:
\begin{align}
    \mathcal{P}_t &= \left\{
    (\kbm^{(t)}_1, \Pbm^{(t)}_1),\dots,(\kbm^{(t)}_M, \Pbm^{(t)}_M) 
    \right\}. \label{eq:prompt_pool}
\end{align}
For prompt selection, we employ the same query-key mechanism as in L2P. We use cosine similarity for the scoring function $\gamma: \mathbb{R}^d \times \mathbb{R}^d \rightarrow \mathbb{R}$ to measure the similarity between the query feature $q(\xbm)$, encoded by a pre-trained BERT model, and the prompt keys. 
Additionally, we introduce a \emph{prompt pool loss}, as in L2P, to facilitate prompt selection:
\begin{align}
    \mathcal{L}_{pp} = \sum_{s \in \boldsymbol{K}_\xbm}
    \gamma(q(\xbm), \kbm_{s}^{(t)}),
\end{align}
where $\boldsymbol{K}_x$ is the set of prompts selected for the instance $\xbm$. 

{However, maintaining a separate prompt pool for each task significantly increases computational costs. To address this issue, we configure each prompt to be a single prefix expert by setting its length to $L = 1$. This approach contrasts with methods that use longer prompts, in which multiple experts must share a single, common key, $\kbm^{(t)}_i$. As explained in Section~\ref{sec:viewpoint}, each prompt $\Pbm_i$ in the pool implicitly encodes a group of $L$ experts, $f_{N + (i-1)L + 1}, \dots, f_{N + iL}$ for $i = 1, \dots, M$. Since each prompt $\Pbm_i$ is tied to a single prompt key $\kbm_i$, all $L$ experts it encodes must share that same key. By setting $L = 1$, we ensure that each expert has its own dedicated key, thereby avoiding the key-sharing mechanism inherent in prior methods such as L2P. This enhances flexibility, as each prefix expert can be selected independently based on the input, leading to a more specialized and adaptive model during both training and inference.}

{Our architecture uses the contextual query feature $q(\xbm)$ to assign distinct sets of experts to different input instances.} This design allows each expert in the prompt pool to {specialize in the specific patterns and data characteristics most relevant to different regions of the input domain}. As a result, the model effectively captures within-task variations. Moreover, by utilizing a task-specific prompt pool, we prevent parameter interference across tasks. {This isolation preserves the integrity of task-specific knowledge, which is crucial for reducing catastrophic forgetting and enhancing overall performance. This mechanism of input-dependent expert selection closely mirrors a SMoE architecture discussed in Section~\ref{sec:background_moe}. For more details, please refer to~\ref{appendix:compare_smoe}.}




\begin{table*}[t]
\small
\centering
\begin{tabular}{@{}p{\linewidth}@{}}
\textbf{Prompt for generate label description} \\
\hline
You are a professional data scientist, working in a relation extraction project. \\
Given a relation and its description, you are asked to write a more detailed description of the relation and provide 3 sentence examples of the relation. \\
The relation is: place served by transport hub \\
The description is: territorial entity or entities served by this transport hub (airport, train station, etc.) \\
Please generate \textbf{2} diverse samples of (relation description, examples). \\
Your response: \\
\hline \\

\textbf{Output:} \\
\hline
This relation identifies the geographic areas or communities that are directly connected to and supported by a specific transport hub, such as an airport, train station, or bus terminal, emphasizing the role of the hub in providing access to transportation services for these areas.\\
Examples: \\
- Heathrow Airport serves the Greater London area, providing access to travelers from London and surrounding towns. \\
- Union Station in Washington, D.C. serves the metropolitan area, facilitating travel for residents and visitors in Maryland and Virginia. \\
- Los Angeles International Airport (LAX) serves not only Los Angeles but also the wider Southern California region, including cities like Santa Monica and Long Beach.\\
\hline 
This relation captures the locations or regions that benefit from the operational capabilities of a transport hub, highlighting its significance in regional connectivity and economic activity.\\
Examples: \\
- Chicago O'Hare International Airport serves the entire Chicago metropolitan area, enhancing accessibility for surrounding suburbs.\\
- Penn Station in New York City is a critical transport hub that serves travelers from New Jersey, New York, and Connecticut.\\
- Dallas/Fort Worth International Airport serves the Dallas and Fort Worth areas, acting as a key transit point for the entire North Texas region.\\
\hline
\end{tabular}
\caption{Example generated label descriptions}
\label{table:judge_prompt_stability}
\vskip -0.1in
\end{table*}

\subsection{Enhancing Training with Label Descriptions} \label{sec:method_label_desc}

{Let $\xbm_e$ be the embeddings of an input sequence $\xbm$. After incorporating prompts via prefix-tuning, the modified input embeddings are denoted as $\xbm_p$.} The model's output is then given by $g_\phi(f_\theta(\xbm_p))$, where $f_\theta$ is the pre-trained transformer encoder and $g_\phi$ is the final relation classifier, parameterized by $\phi$.

In our framework, each task is associated with a distinct prompt pool. At test time, selecting the correct prompt pool for a given sample can be prone to errors. {Such a misalignment may lead to the use of an inappropriate prompt pool, resulting in suboptimal or incorrect predictions.} To mitigate this risk, we aim to improve the generalization capacity of each prompt pool, anticipating that it may encounter samples from tasks it was not originally designed for.

{To achieve this, we propose augmenting the training objective for each prompt pool with representations derived from \emph{label descriptions}.} These representations capture the core attributes of each relation, thereby providing global information that helps reduce overfitting and improve model robustness. Previous studies have demonstrated that label descriptions can significantly benefit CRE tasks by providing additional context on relation types~\citep{yang2020enhance, liu2022simple, borchert2024efficient, luo2024synergistic}. Notably, they offer a consistent, class-specific global perspective, enabling the model to focus on essential relational characteristics and avoid spurious features. 

Specifically, we leverage the detailed definitions provided for each label, typically available in benchmarking datasets, which we refer to as \emph{raw descriptions}. {However, these raw descriptions may lack sufficient diversity and richness, which can introduce noise and instability that harm model performance.} To address these limitations, we follow prior work~\citep{thanh2025few} and employ Gemini 1.5~\citep{team2023gemini, team2024gemini} to generate $D$ diverse, detailed, and illustrative descriptions for each relation. For each label, the raw description is fed into the large language model {to guide the generation process. Our prompt template is shown in Table~\ref{table:judge_prompt_stability}.}

The generated descriptions are then encoded using a pre-trained BERT model to obtain their corresponding description representations. For task $t$, each relation $r \in \mathcal{R}_t$ is associated with a set of label description representations $Des_r = \{\dbm_{r, i} \in \RR^d \ | \ i = 1, \dots, D \}$. {Given an input $\xbm$ with a ground-truth label $y$, our objective is to train prompts that align the input's learned representation, $f_\theta(\xbm_p)$, with the description representations of the correct relation.

To achieve this, we pull $f_\theta(\xbm_p)$ closer to the description representations of the correct label while pushing it away from those of incorrect labels. We formalize this objective using a contrastive loss that enforces separation between the target and irrelevant labels:}
\begin{align}
    \mathcal{L}_{cl} &= - \mathrm{log} 
        \frac
        {
        \sum_{\dbm_y \in Des_y}
        \mathrm{exp}\left(
        f_\theta(\xbm_p) \cdot \dbm_y
        \right)
        }
        {\sum_{r \in \widehat{\mathcal{R}}_t}
        \sum_{\dbm_r \in Des_r}
        \mathrm{exp}\left(
        f_\theta(\xbm_p) \cdot \dbm_r
        \right)
        }, \label{eq:des_loss}
\end{align}
where $Des_y$ is the set of description representations for the ground-truth label $y$. {This loss function encourages high similarity between the input representation and its corresponding label descriptions while minimizing similarity with incorrect ones, thereby enhancing the model's discriminative power.} Consequently, incorporating label descriptions improves the stability and reliability of learned representations, leading to robust performance even when faced with previously unseen or misclassified samples.

\paragraph{Optimization Objective} For each new task $\mathcal{T}_t$, {we train a corresponding prompt pool $\mathcal{P}_t$. During each training step, we select $K$ prompts as previously described}, and the resulting prompted embeddings, $\xbm_p$, are passed to the pre-trained transformer encoder $f_\theta$ and the final classifier $g_\phi$. {The overall objective is to minimize the following loss function:}
\begin{align}
\min_{\mathcal{P}_t, \phi}
\mathcal{L}(g_{\phi}(f_\theta(\xbm_p)), y) 
+ \alpha \mathcal{L}_{pp}
+ \beta \mathcal{L}_{cl}
,\label{eq:prompt-pool-learning-loss}
\end{align}
where $\alpha$ and $\beta$ are hyperparameters. The first term {the standard classification loss. Importantly, when training on task $t$, we only update the parameters of the current prompt pool $\mathcal{P}_t$ and the final classifier $g_{\phi}$.} The pre-trained encoder and all previous prompt pools $\mathcal{P}_1, \dots, \mathcal{P}_{t - 1}$ remain frozen.  

{By combining task-specific prompt pools with label descriptions, our framework \emph{improves within-task prediction performance}, a key component of the overall continual learning objective (see Equation~\eqref{eq:cil_decompostion}). Next, we introduce our strategy to enhance task-identity inference.}

\subsection{Cascade Voting for Task Prediction} \label{sec:method_gen_model}

\begin{algorithm}[tb]
\caption{Cascade Voting Mechanism} \label{alg:cascade_voting}

\textbf{Input}: Voting results $\mathcal{V}^0(\xbm)$ and $\mathcal{V}^1(\xbm)$ of prompt pool $\mathcal{P}_0$ and $\mathcal{P}_1$, Maximum allowed experts $m$, Confidence scores $\{ \text{Score}^i_k(\xbm) \ | \ 2 \leq i \leq t - 1, i \leq k \leq t \}$
\\
\textbf{Output}: Voting result $\mathcal{V}(\xbm)$
\begin{algorithmic}[1] 
\IF{$\mathcal{V}^0(\xbm) = \mathcal{V}^1(\xbm)$}
    \STATE \textbf{return} $\mathcal{V}^0(\xbm)$
\ENDIF
\STATE $start = 0$
\STATE $end = \min(\min(\mathcal{V}^0(\xbm), \mathcal{V}^1(\xbm)), m)$
\STATE result = dict()
\FOR{$i \gets start$ \textbf{to} $end$}
    \STATE {$
\mathcal{V}^i(\xbm)
= \underset{k}{\arg\min} \ \text{Score}^i_k(\xbm), \ k \geq end
$}
    \STATE $\text{result}[\mathcal{V}^i(\xbm)] += 1$
\ENDFOR
\STATE \textbf{return} ${\arg \max_{c}}\  \text{result}[c]$
\end{algorithmic}
\end{algorithm}

{During training, the model is provided with a task-specific dataset and optimizes its corresponding prompt pool.} However, during inference, a key challenge arises: \emph{the model no longer receives explicit task labels for incoming inputs}. Consequently, it must autonomously determine which task-specific prompt pool to employ for generating accurate predictions.

To address this, HiDe-Prompt~\citep{wang2024hierarchical} introduces an additional MLP head $\hat{g}_\psi$, which serves as a task predictor. This head leverages the pre-trained unprompted feature $f_\theta(\xbm)$ to predict the task label.  However, as discussed in Section~\ref{sec:viewpoint}, treating each task as a distinct class can be suboptimal because tasks are often defined by their order of appearance rather than by meaningful semantic differences. In response, \citet{le2024adaptive} increases the predictor’s output dimension to match the number of encountered relations, and then derives the task label from the classified relation. Despite this, this method still relies on pre-trained knowledge and overlooks the potential task-identification capabilities of the specialized experts $\mathcal{P}_t$.

In contrast, our approach leverages Cascade Voting~\citep{zhou2024ensemble} to determine task identity. By discarding the MLP-based classifier head, Cascade Voting eliminates the need to train a task predictor. {Instead, it employs a voting strategy in where the specialized experts in each prompt pool $\mathcal{P}_t$ is collaboratively determine the task identity for a given instance.}

Formally, for an instance $\xbm$ of relation $r$, let $\zbm_r^i = f_{\theta, \mathcal{P}_i}(\xbm) \in \RR^d$ be the representation obtained by passing $\xbm$ through the pre-trained transformer combined with prompt pool $\mathcal{P}_i$. We assume that during the $t$-th task, the class-conditional distribution of $\mathcal{D}^i_{r,t} = \{ \zbm_r^i = f_{\theta, \mathcal{P}_i}(\xbm) | (\xbm, r) \in \mathcal{D}_t \}$ {follows a multivariate Gaussian distribution $\mathcal{N}(\mean^i_{r,t}, \cov^i_t)$ for each prompt pool $\mathcal{P}_i$ and for all relations $r \in \mathcal{R}_t$.} Based on this assumption, we use the maximum likelihood estimator to approximate the mean and covariance as follows:
\begin{align}
    \mean^i_{r,t} &= \frac{1}{|\mathcal{D}^i_{r,t}|} \sum_{\zbm^i_r \in \mathcal{D}^i_{r,t}} \zbm^i_r, 
    \label{eq:feature_mean}
    \\
    \cov^i_t &= \frac{1}{|\mathcal{D}_t|} \sum_{r \in \mathcal{R}_t} \sum_{\zbm^i_r \in \mathcal{D}^i_{r,t}} (\zbm^i_r - \mean^i_{r,t})(\zbm^i_r - \mean^i_{r,t})^\top, \label{eq:feature_covariance}
\end{align}
While each relation–prompt pool pair maintains its own mean vector, a single covariance matrix is shared among all relations within the same task to capture the global variability at the task level. 

{In continual relation extraction, the model is precluded from accessing data from previously encountered tasks. Consequently, it is impossible to establish a distribution for how prompt pool $\mathcal{P}_i$ affects an earlier task $j$ where $j<i$. To overcome it, we employ cascade voting during inference to determine the appropriate prompt pool for relation classification. This approach comprises two stages. First, each prompt pool, viewed as an expert specialized for a particular task, independently assesses and votes on the suitability for the given instance. Second, these votes are aggregated to identify the most suitable prompt pool for that instance.

\begin{figure}[t]
    \centering
    \includegraphics[width=\columnwidth]{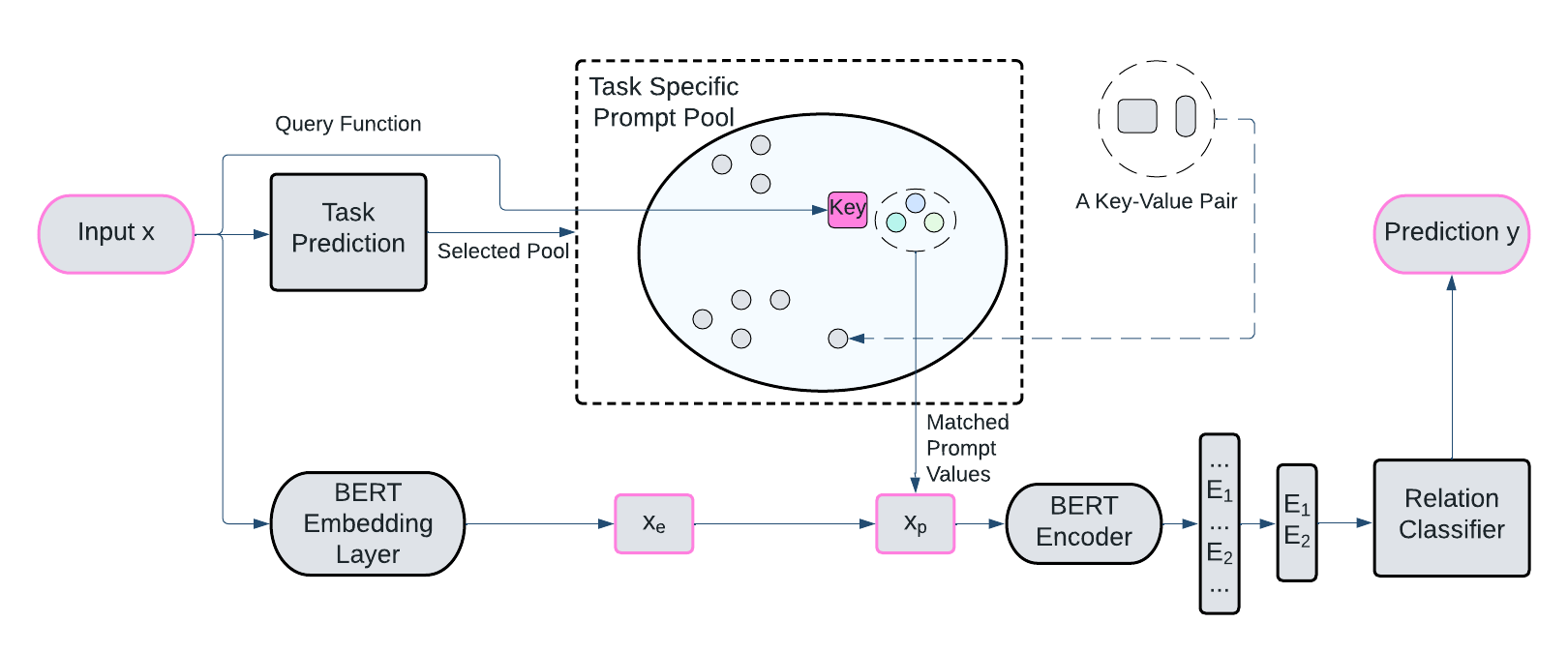}
    \captionsetup{justification=justified, singlelinecheck=false}
    \captionsetup{skip=-7.5pt}
    \caption{Data flow diagram for the inference process. First, the task identity of the input $\xbm$ is inferred via cascade voting to determines the appropriate prompt pool. A query $q(\xbm)$, generated from the input is used to retrieve the prompt from this pool whose key most closely matches the query. The selected prompt is prepended to the input $\xbm_e$, creating the prompted input $\xbm_p$. This sequence is then fed into the BERT encoder. The output embeddings corresponding to the entities $E_1$ and $E_2$ are extracted and concatenated. Finally, the resulting concatenated embedding is passed to the relation classifier $g_\phi$ to predict the relation label $y$.}
    \label{fig:data-flow-diagram}
    \vskip -0.1in
\end{figure}

\paragraph{Voting Procedure for a Prompt Pool} The voting procedure for a prompt pool $\mathcal{P}_i$ begins by generating a feature representation $\zbm^i = f_{\theta, \mathcal{P}_i}(\xbm)$, by passing the input $\xbm$ through the pre-trained backbone $f_\theta$ combined with $\mathcal{P}_i$. Since each relation $r \in \mathcal{R}_t$ is associated with a pre-computed multivariate Gaussian distribution $\mathcal{N}(\mean^i_{r,t},\cov^i_t)$, we calculate a voting score for each task $t \geqslant i$. In particular, the score for task $t$ is defined as the minimum Mahalanobis distance from $\zbm^i$ to the distribution $\{\mathcal{N}(\mean^i_{r,t}, \cov^i_t) \ | \ r \in \mathcal{R}_t \}$ of any relation within that task:
\begin{align}
    \text{Score}^i_t(\xbm) 
    &= \min_{r \in \mathcal{R}_t}
       \bigl(\zbm^i - \mean^i_{r,t}\bigr)^\top
       {\cov^i_t}^{-1}
       \bigl(\zbm^i - \mean^i_{r,t}\bigr),
       \quad t \geqslant i.
\end{align}
Finally, the vote cast by $\mathcal{P}_i$  is the task identity that minimizes this score:
\begin{align}
    \mathcal{V}^i(\xbm) = {\arg \min_{t}}\  \text{Score}^i_t(\xbm), \quad  
    t \geqslant i. \label{eq:voting}
\end{align}

\paragraph{Cascade Voting Mechanism} Since each prompt pool's valid prediction ranges can vary, we employ a cascade voting mechanism to mitigate these boundary mismatches. The pre-trained BERT model, $\mathcal{P}_0$, serves as the initial expert. We first consider the votes from $\mathcal{P}_0$ and $\mathcal{P}_1$, as both are capable of predicting any subsequent task. If $\mathcal{P}_0$ and $\mathcal{P}_1$ concur on a task label, that label is immediately assigned to $\xbm$.

{If $\mathcal{P}_0$ and $\mathcal{P}_1$ yield conflicting predictions, we resolve the disagreement by aggregating votes from multiple experts. An expert $\mathcal{P}_i$ is permitted to predict task identities $j \geq i$. To maintain computational efficiency and a consistent number of voting experts, we define an index interval $[start, end]$ specifying the experts eligible to participate. For fairness, all experts within this interval make predictions over the task identity range $[end, t]$, where $t$ denotes the index of the current task. Since $\mathcal{P}_0$ can predict all task identities, we set $start = 0$. To determine the upper bound $end$ while treating $\mathcal{P}_0$ and $\mathcal{P}_1$ symmetrically, we choose the smaller of their predicted task identities, ensuring that both predictions are included in the voting process. To prevent an excessive number of voting experts, we impose a threshold $m$ on the upper bound, and thus set
$end = \min(\min (\mathcal{V}^0, \mathcal{V}^1), m)$.}

The task identity that receives the majority of votes is then assigned to $\xbm$, and the corresponding prompt pool is selected for relation classification. This cascade voting strategy thus limits the number of prompt pools involved at each decision step and effectively handles inconsistencies in prediction ranges. A detailed algorithmic description is provided in Algorithm~\ref{alg:cascade_voting}. As demonstrated in Section~\ref{sec:exp_ablation}, our approach enhances task-identity prediction accuracy. This improvement boosts TII performance, which, according to Equation~\eqref{eq:cil_decompostion}, leads to more effective continual relation extraction. For a comparison with the MLP-based approach, see~\ref{appendix:cascade_mlp}.}

\subsection{Mitigating Classification Bias with Generative Models} \label{sec:method_tap}

A key element of our framework is the MLP head relation classifier $g_\phi$. To mitigate catastrophic forgetting, we retain the representations of previously learned relations by deploying a generator that encodes their input distribution. {Specifically, for each relation $r \in \mathcal{R}_t$, we use a Gaussian distribution $\mathcal{N}(\mean^t_{r,t},\cov^t_t)$. The mean and covariance are computed from the prompt pool $\mathcal{P}_t$ of the current task, as described in Section~\ref{sec:method_gen_model}.}

\begin{algorithm}[tb]
\caption{$\mathcal{T}_t$ training process} \label{alg:cap}

\textbf{Input}: Training $t$-th dataset $\mathcal{D}_t$, relation set $\mathcal{R}_t$\\
\textbf{Output}: Prompt pool $\mathcal{P}_t$ and relation classifier $\phi$
\begin{algorithmic}[1] 
\STATE Randomly initialize $\mathcal{P}_t$
\FOR{$epoch \gets 1$ \textbf{to} $training\_epoch$}
    \FOR{batch $\xbm_{B} \subset \mathcal{D}_t$}
        \STATE Update $\mathcal{P}_t$ and $g_{\phi}$ on $\xbm_{B}$ via Equation~\eqref{eq:prompt-pool-learning-loss}
    \ENDFOR
\ENDFOR
\STATE Update $\hat{\mathcal{R}}_{t} \gets \hat{\mathcal{R}}_{t-1} \cup \mathcal{R}_t$

\FOR{each $r \in \mathcal{R}_t$}
    \FOR{$i \gets 1$ \textbf{to} $t$}
        \STATE Compute $\mean^i_{r, t}$ and $\cov^i_t$ via Equation~\eqref{eq:feature_mean} and Equation~\eqref{eq:feature_covariance}
    \ENDFOR
\ENDFOR
\STATE Train the relation classifier $\phi$ via Equation~\eqref{eq:tap}
\STATE \textbf{return} $\mathcal{P}_{t},  \phi$
\end{algorithmic}
\end{algorithm}

By using a prompt-based generative model, the representations of previously learned relations can be reconstructed and replayed for subsequent tasks, thereby preserving essential relational knowledge. {Storage requirements for each relation are minimal, as only a mean vector and a covariance matrix are stored. Furthermore, this design effectively addresses catastrophic forgetting by eliminating the need to store observed instances explicitly. Instead, the generative models capture and reproduce the distributional characteristics of each relation's representations, enabling the model to simulate previously encountered relations without directly accessing historical data.} This approach not only enhances computational efficiency but also ensures compliance with privacy constraints, as raw data is never retained.

These generative models produce synthetic representations $\zbm_r$ for each relation $r$, which are then used to train the relation classifier.
The MLP classifier $g_\phi$ is trained by minimizing the following cross-entropy loss:
\begin{align} \label{eq:tap}
        \mathcal{L}(\phi) = 
        \sum_{i = 1}^t
        \sum_{r \in \mathcal{R}_i} \sum_{\zbm_r \sim \mathcal{N}(\mean^t_{r,t},\cov^t_t)}
        -\mathrm{log} \frac
        {\exp(g_\phi(\zbm_r)[r])}
        {\sum_{r' \in \widehat{\mathcal{R}}_t} \exp(g_\phi(\zbm_r)[r'])}.
\end{align}
The complete training procedure is detailed in Algorithm~\ref{alg:cap}, and the data flow for the inference process is illustrated in Figure~\ref{fig:data-flow-diagram}.

\begin{table}[!t]
\vskip -0.2in
    \centering
    \begin{tabular}{l | c c c c c c c c c c}
    \hline
    \multicolumn{11}{c}{\textbf{FewRel}} \\
    \hline
    Model  & $\mathcal{T}_1$ & $\mathcal{T}_2$ & $\mathcal{T}_3$ & $\mathcal{T}_4$ & $\mathcal{T}_5$ & $\mathcal{T}_6$ & $\mathcal{T}_7$ & $\mathcal{T}_8$ & $\mathcal{T}_9$ & $\mathcal{T}_{10}$ \\
    \hline
    RP-CRE & 97.9 & 92.7 & 91.6 & 89.2 & 88.4 & 86.8 & 85.1 & 84.1 & 82.2 & 81.5 \\
    ACA & 98.3 & 95.0 & 92.6 & 91.3 & 90.4 & 89.2 & 87.6 & 87.0 & 86.3 & 84.7 \\
    CRL & 98.1 & 94.6 & 92.5 & 90.5 & 89.4 & 87.9 & 86.9 & 85.6 & 84.5 & 83.1 \\
    CDec & 98.4 & 95.4 & 93.2 & 92.1 & 91.0 & 89.7 & 88.3 & \underline{87.4} & \underline{86.4} & 84.8 \\
    CEAR & 98.1 & \underline{95.8} & 93.6 & 91.9 & 91.1 & 89.4 & 88.1 & 86.9 & 85.6 & 84.2 \\
    RationaleCL & 98.6 & 95.7 & 93.4 & \underline{92.3} & \underline{91.3} & 89.7 & 88.2 & 87.3 & 86.3 & \underline{85.1} \\
    CREST & \underline{98.7} & 93.6 & \underline{93.8} & \underline{92.3} & 91.0 & \underline{89.9} & 87.6 & 86.7 & 86.0 & 84.8 \\
    DP-CRE & 98.5 & 95.4 & 93.7 & 92.1 & 90.9 & 89.4 & \underline{88.5} & \underline{87.4} & 86.3 & \underline{85.1} \\
    \hline
    L2P & 97.4 & 90.8 &	83.6 &	76.5 &	68.9 &	64.1 &	61.0 &	57.4 &	50.1 &	44.6 \\
    HiDe-Prompt & 95.5 &	89.4 &	86.0 &	85.7 &	87.8 &	84.2 &	75.9 &	75.1 &	70.3 &	67.2 \\
    EoE & 97.8 & 95.0 & 93.6 & 92.5 & 91.6 & 90.0 & 88.9 & 87.9 & 86.9 & 85.5 \\
    WAVE-CRE & 97.9 &	95.5 &	93.6 &	92.4 &	91.1 &	90.2 &	88.7 &	87.6 &	86.5 &	85.0 \\
    \hline
    WAVE++  &\textbf{98.2} &	\textbf{95.8} &	\textbf{95.1}   &	\textbf{94.1} & \textbf{92.7}   &	\textbf{91.9}   &	\textbf{90.2} &	\textbf{89.9} &	\textbf{89.0} &	\textbf{87.7} \\
    \hline
    \hline
    \multicolumn{11}{c}{\textbf{TACRED}} \\
    \hline
    Model  & $\mathcal{T}_1$ & $\mathcal{T}_2$ & $\mathcal{T}_3$ & $\mathcal{T}_4$ & $\mathcal{T}_5$ & $\mathcal{T}_6$ & $\mathcal{T}_7$ & $\mathcal{T}_8$ & $\mathcal{T}_9$ & $\mathcal{T}_{10}$ \\
    \hline    
    RP-CRE & 97.6 & 90.6 & 86.1 & 82.4 & 79.8 & 77.2 & 75.1 & 73.7 & 72.4 & 72.4 \\
    ACA & 98.0 & 92.1 & 90.6 & 85.5 & 84.4 & 82.2 & 80.0 & 78.6 & 78.8 & 78.1 \\
    CRL & 97.7 & 93.2 & 89.8 & 84.7 & 84.1 & 81.3 & 80.2 & 79.1 & 79.0 & 78.0 \\
    CDec & 97.7 & 92.8 & 91.0 & 86.7 & 85.2 & 82.9 & 80.8 & 80.2 & 78.8 & 78.6 \\
    CEAR & 97.7 & 94.3 & \underline{92.3} & \underline{88.4} & \underline{86.6} & 84.5 & 82.2 & 81.1 & 80.1 & 79.1 \\
    RationaleCL & \underline{98.6} & \underline{94.4} & 91.5 & 88.1 & 86.5 & \underline{84.9} & \underline{84.5} & \underline{82.5} & \underline{81.6} & \underline{80.8} \\
    CREST & 97.3 & 91.4 & 82.3 & 82.5 & 79.2 & 75.8 & 78.8 & 77.4 & 78.6 & 79.4 \\
    DP-CRE & 97.8 & 93.8 & 91.5 & 87.5 & 85.7 & 84.2 & 82.9 & 81.3 & 81.5 & 80.7 \\
    \hline
    L2P & 96.9 & 88.2 &	73.8 &	68.6 &	66.3 &	63.1 &	60.4 &	59.1 &	56.8 &	54.8 \\
    HiDe-Prompt	& 97.3 &	92.8 &	86.2 &	82.6 &	80.6 &	80.4 &	75.8 &	73.7 &	72.9 &	72.6 \\
    EoE & \textbf{98.7} & \textbf{94.7} & 90.6 & 87.8 & \textbf{87.2} & \textbf{85.9} & \textbf{84.3} & 83.2 & 82.7 & 81.5 \\
    WAVE-CRE &	98.4 &	94.3 &	\textbf{91.6} &	87.8 &	85.7 &	83.5 &	81.3 &	80.4 &	79.5 &	78.7 \\
    \hline
    WAVE++ &	97.6 &	93.6 &	90.7 &	\textbf{88.2} & 86.4 & 85.4 & \textbf{84.3} &\textbf{83.7} & \textbf{83.2} &	\textbf{82.5}\\
    \hline
    \end{tabular}
    \caption{Average accuracy (\%) of all methods across learning stages for FewRel and TACRED dataset. The best accuracy scores under the rehearsal-free and rehearsal-based setting are in \textbf{bold} and \underline{underlined}, respectively.}
    \label{tab: acc}
\vskip -0.1in
\end{table} 

\section{Experiments} \label{sec:experiments}

\subsection{Experimental Settings}

\paragraph{Datasets} To ensure consistency and fairness in our evaluation, we follow the protocols established by \citet{zhou2024ensemble, le2024adaptive} and assess the performance of WAVE++ on two widely used CRE datasets: 
\begin{itemize} 
    \item \textbf{TACRED} \citep{zhang2017position} comprises 106,264 instances across 41 relation types. Following the experimental setup of \citet{cui2021refining}, we partition this dataset into 10 distinct subsets. The number of training samples of each relation is limited to 320, while the number of test samples of each relation is limited to 40.
    \item \textbf{FewRel} \citep{han2018fewrel} contains 56,000 instances across 80 relation types. Following the setup of \citet{wang2019sentence}, we divide this dataset into 10 distinct subsets. 
\end{itemize}

\paragraph{Baseline Models} We compare WAVE++ with several prompt-based methods, including L2P~\citep{wang2022learning}, HiDe-Prompt \citep{wang2024hierarchical}, EoE \citep{zhou2024ensemble} and WAVE-CRE \citep{le2024adaptive}. Notably, WAVE-CRE \citep{le2024adaptive} is a simplified version of our proposed WAVE++. While WAVE-CRE also employs task-specific prompt pools, WAVE++ extends this strategy by integrating label descriptions to better capture relational information, and replacing the MLP classifier with cascade voting for task identity prediction. We further contrast our approach with traditional rehearsal-based methods: RP-CRE \citep{cui2021refining}, ACA \citep{wang2022learning}, CRL \citep{zhao2022consistent}, CDec \citep{xia2023enhancing}, CEAR \citep{zhao2023improving}, RationaleCL \citep{xiong2023rationale}, CREST \citep{le2024continual}, and DP-CRE \citep{huang2024dp}. For all methods, we utilize a pre-trained BERT encoder \citep{devlin2018bert} as the backbone architecture. We report model performance in terms of average accuracy across five distinct random runs.

\paragraph{Implementation Details} All experiments were conducted on a single NVIDIA A100 GPU. We tuned the hyperparameters for WAVE++ using random search, while for baseline methods, we adopted the configurations reported in \citet{zhao2022consistent} to ensure a fair comparison. Specifically, we froze the parameters of the pre-trained BERT model and optimized only the MLP-based relation classifiers and the prompt pools. WAVE++ has 3.5 million trainable parameters, approximately 300,000 fewer than the original WAVE model (3.8 million). In practice, the model trains for approximately seven hours on FewRel and one hour on TACRED. {Following \citet{zhou2024ensemble}, the maximum number of experts allowed in cascade voting is set to $m = 2$. A detailed analysis of training and inference time is provided in~\ref{appendix:time_analysis}.}

\subsection{Main Results}

In Table~\ref{tab: acc}, we present a detailed performance comparison of WAVE++ against both rehearsal-free and rehearsal-based CRE methods. We first benchmark WAVE++ against other rehearsal-free approaches and find that it consistently outperforms all baselines. Notably, on the final task of the FewRel dataset, WAVE++ exceeds the performance of the second-best method, EoE, by 2.0\%. {This superiority is consistent across various training stages, as demonstrated in Figure~\ref{fig:task_acc_fewrel}. Although WAVE++ exhibits minor performance dips relative to EoE at certain points on TACRED, its strong performance on the final task highlights its superior resistance to catastrophic forgetting.}

\begin{figure}[!t]
\begin{center}
\centerline{\includegraphics[width=0.9\columnwidth]{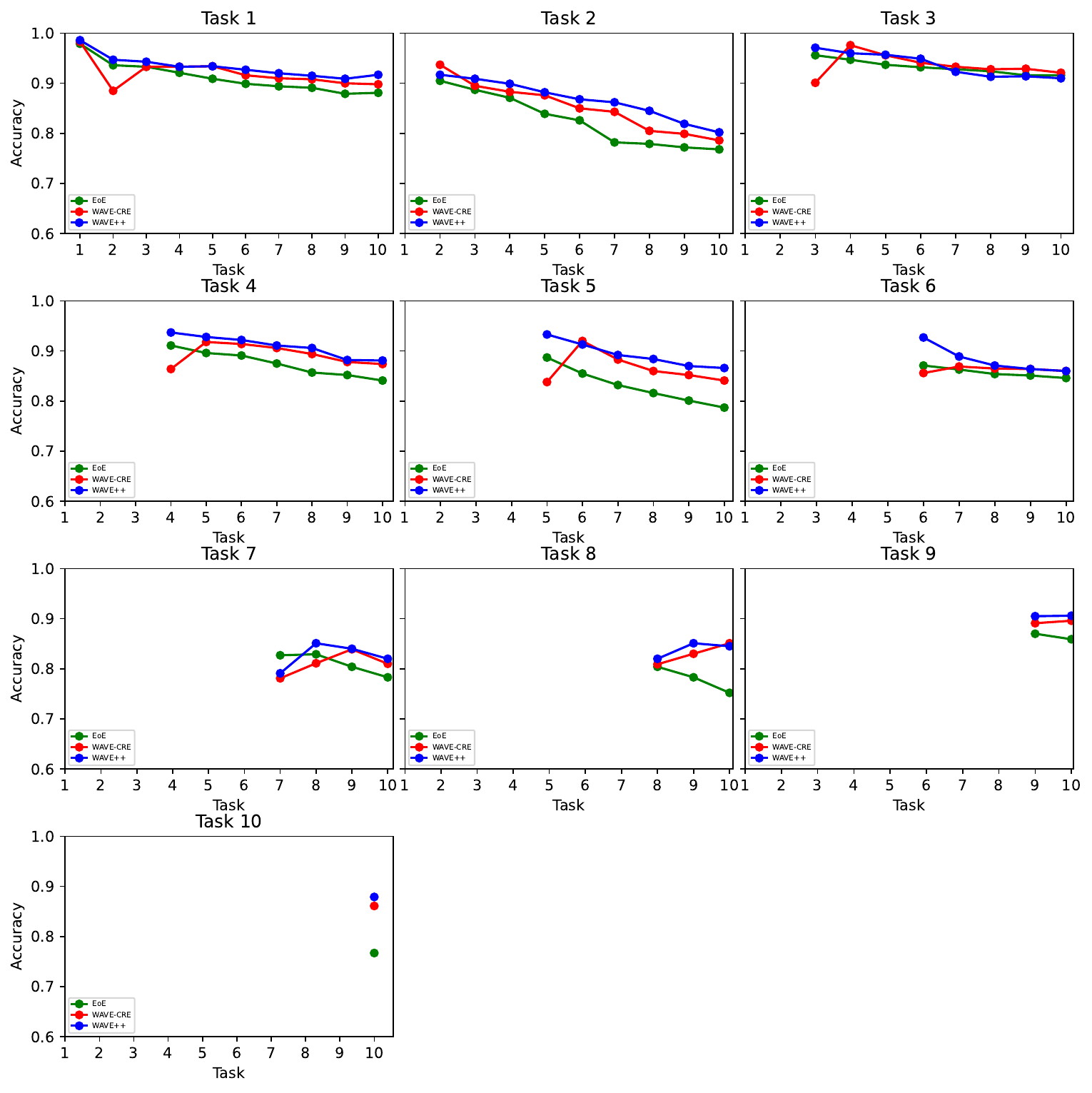}}
\caption{Average accuracy (\%) on individual tasks throughout the continual learning process for WAVE++, WAVE-CRE, and EoE on the FewRel dataset.}
\label{fig:task_acc_fewrel}
\end{center}
\vskip -0.4in
\end{figure}

 WAVE++ also demonstrates substantial improvements over its predecessor, WAVE-CRE, throughout the training process. By the final task, WAVE++ achieves an accuracy that surpasses WAVE-CRE by approximately 2.0\% on FewRel and 4.0\% on TACRED. The accuracy across individual tasks also shows significant improvement over the course of training, as depicted in Figure~\ref{fig:task_acc_fewrel}. These gains can largely be attributed to the incorporation of label descriptions and an enhanced voting-based task prediction mechanism, which replaces the previous MLP-based approach. {A detailed ablation study of these components is presented in Section~\ref{sec:exp_ablation}.}

Furthermore, we evaluate WAVE++ against recent state-of-the-art, rehearsal-based CRE methods. Despite not utilizing a memory buffer for replay, WAVE++ outperforms all evaluated rehearsal-based methods. On both datasets, WAVE++ demonstrates a significant performance advantage and exhibits robust resistance to catastrophic forgetting, all without storing prior samples. These findings underscore the effectiveness and efficiency of WAVE++ as a powerful solution for continual relation extraction.

\subsection{Ablation Studies} \label{sec:exp_ablation}

{\paragraph{Effect of Task-Specific Prompt Pool} To assess the impact of the task-specific prompt pool, we compare the full WAVE++ model against an ablated variant restricted to a single prompt per task. The results, presented in Table~\ref{table:wave_wo_prompt_pool}, show that using a prompt pool consistently enhances performance across both datasets.} As discussed in Section~\ref{sec:method_prompt_pool}, the prompt pool enables the dynamic selection of prefix experts within each task, rather than relying on a single, static set. This flexibility allows the model to better capture task-specific variations, leading to improved performance.

\begin{table}[t]
    \centering
    \begin{tabular}{l|c c c c c c c c c c}
    \hline
    \multicolumn{11}{c}{\textbf{FewRel}} \\
    \hline
    Model  & $\mathcal{T}_1$ & $\mathcal{T}_2$ & $\mathcal{T}_3$ & $\mathcal{T}_4$ & $\mathcal{T}_5$ & $\mathcal{T}_6$ & $\mathcal{T}_7$ & $\mathcal{T}_8$ & $\mathcal{T}_9$ & $\mathcal{T}_{10}$ \\
    \hline 
    WAVE++ w/o Prompt Pool& 96.1 & 93.9 & 92.6 & 92.4 &	90.3 &	89.4 &	88.5 &	87.9 &	87.9 &	86.4 \\
    WAVE++	& \textbf{98.2} &	\textbf{95.8} &	\textbf{95.1}   &	\textbf{94.1} & \textbf{92.7}   &	\textbf{91.9}   &	\textbf{90.2} &	\textbf{89.9} &	\textbf{89.0} &	\textbf{87.7} \\
    \hline
    \hline
    \multicolumn{11}{c}{\textbf{TACRED}} \\
    \hline
    Model  & $\mathcal{T}_1$ & $\mathcal{T}_2$ & $\mathcal{T}_3$ & $\mathcal{T}_4$ & $\mathcal{T}_5$ & $\mathcal{T}_6$ & $\mathcal{T}_7$ & $\mathcal{T}_8$ & $\mathcal{T}_9$ & $\mathcal{T}_{10}$ \\
    \hline 
    WAVE++ w/o Prompt Pool & 97.5 &	92.8 &	89.1 &	85.9 &	84.4 &	83.7 &	83.1 &	81.9 &	81.6 &	81.1 \\
    WAVE++ & \textbf{97.6} &	\textbf{93.6} &	\textbf{90.7} &	\textbf{88.2} & \textbf{86.4} & \textbf{85.4} & \textbf{84.3} &\textbf{83.7} & \textbf{83.2} &	\textbf{82.5} \\
    \hline
    \end{tabular}
    \caption{Average accuracy (\%) of WAVE++ with and without prompt pool. The best result for each dataset and metric is \textbf{bolded}.} \label{table:wave_wo_prompt_pool}
\end{table}
\begin{table}[t]
    \centering
    \begin{tabular}{l|c c c c c c c c c c}
    \hline
    \multicolumn{11}{c}{\textbf{FewRel}} \\
    \hline
    Model  & $\mathcal{T}_1$ & $\mathcal{T}_2$ & $\mathcal{T}_3$ & $\mathcal{T}_4$ & $\mathcal{T}_5$ & $\mathcal{T}_6$ & $\mathcal{T}_7$ & $\mathcal{T}_8$ & $\mathcal{T}_9$ & $\mathcal{T}_{10}$ \\
    \hline 
    WAVE++ w/o Description	& 96.9 & 94.0 & 92.6 & 91.7 &	90.9 &	89.8 &	88.7 &	88.2 &	87.2 &	85.8 \\
    WAVE++	& \textbf{98.2} &	\textbf{95.8} &	\textbf{95.1}   &	\textbf{94.1} & \textbf{92.7}   &	\textbf{91.9}   &	\textbf{90.2} &	\textbf{89.9} &	\textbf{89.0} &	\textbf{87.7} \\
    \hline
    \hline
    \multicolumn{11}{c}{\textbf{TACRED}} \\
    \hline
    Model  & $\mathcal{T}_1$ & $\mathcal{T}_2$ & $\mathcal{T}_3$ & $\mathcal{T}_4$ & $\mathcal{T}_5$ & $\mathcal{T}_6$ & $\mathcal{T}_7$ & $\mathcal{T}_8$ & $\mathcal{T}_9$ & $\mathcal{T}_{10}$ \\
    \hline 
    WAVE++ w/o descriptions & 97.4 &	93.0 &	89.1 &	86.0 &	84.4 &	83.5 &	82.8 &	81.8 &	81.1 &	80.7 \\
    WAVE++ & \textbf{97.6} &	\textbf{93.6} &	\textbf{90.7} &	\textbf{88.2} & \textbf{86.4} & \textbf{85.4} & \textbf{84.3} &\textbf{83.7} & \textbf{83.2} &	\textbf{82.5} \\
    \hline
    \end{tabular}
    \caption{Average accuracy (\%) of WAVE++ with and without label descriptions. The best result for each dataset and metric is \textbf{bolded}}
    \label{tab:wave_des}
\end{table}
\begin{table}[!t]
    \centering
    \begin{tabular}{l|c c c c c c c c c c}
    \hline
    \multicolumn{11}{c}{\textbf{TACRED}} \\
    \hline
    Model  & $\mathcal{T}_1$ & $\mathcal{T}_2$ & $\mathcal{T}_3$ & $\mathcal{T}_4$ & $\mathcal{T}_5$ & $\mathcal{T}_6$ & $\mathcal{T}_7$ & $\mathcal{T}_8$ & $\mathcal{T}_9$ & $\mathcal{T}_{10}$ \\
    \hline 
    EoE &	98.7 & 94.7 & 90.6 & 87.8 & 87.2 & 85.9 & 84.3 & 83.2 & 82.7 & 81.5 \\
    EoE + descriptions &	\textbf{98.8} &	\textbf{95.3} &	\textbf{91.3} &	\textbf{88.0} &	\textbf{87.5} & \textbf{86.1} & \textbf{84.8} &	\textbf{83.6} & \textbf{83.0} & \textbf{81.9} \\
    \hline
    \end{tabular}
    \caption{Average accuracy (\%) of EoE with and without label descriptions. The best result for each dataset and metric is \textbf{bolded}}
    \label{tab:eoe_des}
\vskip -0.1in
\end{table}

\paragraph{Effect of Label Descriptions} To comprehensively evaluate the impact of label descriptions, we evaluate both our proposed WAVE++ model and the EoE baseline with and without this component. As shown in Table~\ref{tab:wave_des}, removing label descriptions reduces the average accuracy of WAVE++ by approximately 2.0\% across both datasets.This performance gain can be attributed to the additional contextual information provided by the descriptions, which enables the model to learn more robust relational features and, consequently, enhances classification accuracy. These findings underscore the crucial role of label descriptions in improving within-task prediction performance.

Furthermore, we observe that label descriptions also significantly improve the performance of the EoE baseline on the TACRED dataset, as detailed in Table~\ref{tab:eoe_des}. This result confirms that integrating label descriptions is a broadly effective strategy for enhancing performance in continual relation extraction. A further ablation on the impact of the number of descriptions is provided in~\ref{appendix:num_desc}.

\paragraph{Effect of Number of Experts per Prompt $L$} We also evaluated the impact of the prompt length $L$ on model performance. To ensure a fair comparison, the number of selected prompts $K$, was adjusted to keep the total number of experts constant across all experiments. The results, summarized in Table~\ref{tab: prompt_size}. As discussed in Section~\ref{sec:method_prompt_pool}, setting $L = 1$ maximizes selection flexibility by assigning each prefix expert a unique key, thereby enhancing the model's expressive power. Empirically, this configuration yields the best performance compared to all other tested values of $L$.

\paragraph{Effectiveness of Cascade Voting for Task Identity Inference} 
We replace the conventional MLP head for task classification with a cascade voting mechanism for task identity prediction. This approach both reduces training overhead and improves task identification by aggregating predictions from all prompt pools. To evaluate its effectiveness, we compare it against the MLP-based method used in WAVE-CRE~\citep{le2024adaptive}. The task classification accuracies are presented in Table~\ref{tab: taskid}.

\begin{table}[!t]
    \centering
    \begin{tabular}{l | c c c c c c c c c c}
    \hline
    \multicolumn{11}{c}{\textbf{TACRED}} \\
    \hline
    Settings  & $\mathcal{T}_1$ & $\mathcal{T}_2$ & $\mathcal{T}_3$ & $\mathcal{T}_4$ & $\mathcal{T}_5$ & $\mathcal{T}_6$ & $\mathcal{T}_7$ & $\mathcal{T}_8$ & $\mathcal{T}_9$ & $\mathcal{T}_{10}$ \\
    \hline
$L$=8, $K$=1 & 97.3 & 93.3 & 90.2 & 87.9 & 86.0 & 84.9 & 84.1 & 83.2 & 82.7 & 82.3 \\
$L$=4, $K$=2 & 97.3 & 93.4 & 90.3 & 88.0 & 86.2 & 85.1 & 84.2 & 83.3 & 82.8 & 82.2 \\
$L$=2, $K$=4 & 97.5 & 93.4 & 90.5 & 88.1 & 86.2 & 85.2 & 84.0 & 83.5 & 82.7 & 82.1 \\
$L$=1, $K$=8 & \textbf{97.6} & \textbf{93.6} & \textbf{90.7} & \textbf{88.2} & \textbf{86.4} & \textbf{85.4} & \textbf{84.3} & \textbf{83.7} & \textbf{83.2} & \textbf{82.5} \\
    \hline
    \end{tabular}
    \caption{Detailed analyses of the number of experts $L$ within a prompt. The best result is \textbf{bolded}.}
    \label{tab: prompt_size}
\end{table}
\begin{table}[!t]
    \centering
    \begin{tabular}{l | c c c c c c c c c c}
    \hline
    \multicolumn{11}{c}{\textbf{FewRel}} \\
    \hline
    Model  & $\mathcal{T}_1$ & $\mathcal{T}_2$ & $\mathcal{T}_3$ & $\mathcal{T}_4$ & $\mathcal{T}_5$ & $\mathcal{T}_6$ & $\mathcal{T}_7$ & $\mathcal{T}_8$ & $\mathcal{T}_9$ & $\mathcal{T}_{10}$ \\
    \hline
    EoE & 100 & 95.3 & 95.7 & 94.4 & 92.5 & 91.6 & 88.3 & 87.5 & 86.8 & 86.0 \\
    WAVE-CRE & 100 & 96.8 & 94.5 & 93.3 & 91.0 & 89.5 & 88.6 & 87.6 & 86.4 & 85.4 \\
    WAVE++ & 100 & \textbf{97.9} & \textbf{96.1} & \textbf{95.1} & \textbf{94.0} & \textbf{92.7} & \textbf{91.7} & \textbf{90.8} & \textbf{89.6} & \textbf{88.3} \\
    \hline
    \hline
    \multicolumn{11}{c}{\textbf{TACRED}} \\
    \hline
    Model  & $\mathcal{T}_1$ & $\mathcal{T}_2$ & $\mathcal{T}_3$ & $\mathcal{T}_4$ & $\mathcal{T}_5$ & $\mathcal{T}_6$ & $\mathcal{T}_7$ & $\mathcal{T}_8$ & $\mathcal{T}_9$ & $\mathcal{T}_{10}$ \\
    \hline    
    EoE & 100 & 95.2 & 87.5 & 85.0 & 82.0 & 82.4 & 83.3 & 82.7 & 84.1 & 82.4 \\
    WAVE-CRE & 100 & \textbf{96.8} & 91.1 & 86.8 & 84.9 & 83.6 & 82.1 & 80.4 & 80.2 & 79.2 \\
    WAVE++ & 100 & 95.9 & \textbf{92.0} & \textbf{90.5} & \textbf{88.7} & \textbf{87.9} & \textbf{86.9} & \textbf{85.9} & \textbf{85.2} & \textbf{84.8} \\
    \hline
    \end{tabular}
    \caption{Task prediction accuracy  (\%) of WAVE-CRE and WAVE++. The best result for each dataset and metric is \textbf{bolded}.}
    \label{tab: taskid}
\end{table}
{\begin{table}[!t]
    \centering
{\begin{tabular}{l|c c c c c c c c c c}
\hline
\multicolumn{11}{c}{\textbf{FewRel}} \\
\hline
Model  & $\mathcal{T}_1$ & $\mathcal{T}_2$ & $\mathcal{T}_3$ & $\mathcal{T}_4$ & $\mathcal{T}_5$ & $\mathcal{T}_6$ & $\mathcal{T}_7$ & $\mathcal{T}_8$ & $\mathcal{T}_9$ & $\mathcal{T}_{10}$ \\
\hline 
WAVE++ w/o GP	& 98.2 & 92.2 & 88.5 & 84.2 &	80.2 &	76.2 &	72.9 &	70.2 &	67.3 &	62.1 \\
WAVE++	& \textbf{98.2} &	\textbf{95.8} &	\textbf{95.1}   &	\textbf{94.1} & \textbf{92.7}   &	\textbf{91.9}   &	\textbf{90.2} &	\textbf{89.9} &	\textbf{89.0} &	\textbf{87.7} \\
\hline
\hline
\multicolumn{11}{c}{\textbf{TACRED}} \\
\hline
Model  & $\mathcal{T}_1$ & $\mathcal{T}_2$ & $\mathcal{T}_3$ & $\mathcal{T}_4$ & $\mathcal{T}_5$ & $\mathcal{T}_6$ & $\mathcal{T}_7$ & $\mathcal{T}_8$ & $\mathcal{T}_9$ & $\mathcal{T}_{10}$ \\
\hline 
WAVE++ w/o GP & 97.6 &	93.6 &	80.3 &	72.2 &	70.1 &	67.2 &	65.1 &	63.6 &	62.1 &	60.3 \\
WAVE++ & \textbf{97.6} &	\textbf{93.6} &	\textbf{90.7} &	\textbf{88.2} & \textbf{86.4} & \textbf{85.4} & \textbf{84.3} &\textbf{83.7} & \textbf{83.2} &	\textbf{82.5} \\
\hline
\end{tabular}}
    \caption{Average accuracy (\%) of WAVE++ with and without generative replay. The best result for each dataset and metric is \textbf{bolded}}
    \label{tab: gpr}
    \vskip -0.1in
\end{table}}

The results demonstrate that WAVE++ significantly outperforms WAVE-CRE, achieving an accuracy improvement of approximately 3.0\% on FewRel and 4.0\% on TACRED after the final task. Beyond these accuracy gains, the cascade voting mechanism exhibits superior stability. The MLP-based approach, in contrast, shows occasional sharp declines in accuracy, indicating a sensitivity to distribution shifts when new tasks are introduced. The voting mechanism, however, maintains consistent performance, highlighting its robustness to heterogeneous task distributions. Moreover, WAVE++ also surpasses EoE in task prediction accuracy. These findings underscore the effectiveness and resilience of the cascade voting mechanism in complex, multi-task scenarios where robustness to distribution shifts is crucial.

{\paragraph{Effect of Generative Replay}
To assess the impact of generative replay, we remove it from the WAVE++ MLP relation classifier (Table~\ref{tab: gpr}). The exclusion of generative replay leads to a substantial performance decline across all tasks, a clear indicator of catastrophic forgetting. By the final training stage, overall performance falls by more than 22\%. On the FewRel dataset, for instance, accuracy on previously learned tasks drops by up to 4\% after the introduction of each new task. Similarly, a sharp decline is observed on the TACRED dataset, where performance drops by 13\% between the second and third tasks. These results underscore the critical role of generative replay in mitigating forgetting and preserving knowledge.}

\section{Conclusion}

In this paper, we introduce a novel perspective on prompt-based continual learning by interpreting existing methods through the lens of mixture of experts models. We demonstrated that contemporary prompt-based continual learning approaches can be viewed as a special case of MoE architectures. Building on this insight, we present WAVE++, which shares key conceptual similarities with sparse MoE designs. Our framework addresses several challenges in prompt-based continual learning, including suboptimal task identification, sensitivity to distribution shifts, and difficulties in handling within-task diversity. Specifically, we mitigate these issues by employing a cascade voting mechanism for task prediction and using task-specific prompt pools, which are complemented by label descriptions to enhance continual relation extraction performance.
Despite these advancements, hurdles remain in improving the efficiency and effectiveness of WAVE++. Notably, the method heavily depends on the design of prompt pools and experts, suggesting opportunities to refine and optimize these components. While our approach attenuates catastrophic forgetting, retaining knowledge of previous tasks remains challenging, echoing issues highlighted in prior studies. In particular, if prompt pools are not properly applied during testing, forgetting can reemerge. Moreover, although the prompt pools enhance the expressiveness of our method, the current prefix-tuning experts are relatively simplistic. Future work may explore more sophisticated expert architectures to further enhance the model's capabilities.


\clearpage

\bibliographystyle{cas-model2-names}
\bibliography{references}
%
%
\newpage
\appendix

\section{Proof of Propositions}

In this appendix, we provide proofs for key propositions in the main text.

\subsection{Proof of Proposition~\ref{props: attn_moe}} \label{appendix:proof_of_attn_moe}

Specifically, let $\Xbm = \left[\xbm_1^\top,\dots,\xbm_N^\top\right]^\top \in \RR^{Nd}$ denote the concatenation of all input token embeddings. We define $E_j \in \RR^{d \times Nd}$ as a selector matrix such that $E_{j} \Xbm = \xbm_{j}$, thereby extracting the $j$-th token from the input sequence. From Equation~\eqref{eq:msa}, the output of the $l$-th attention head can be expressed as follows:
\begin{align}
    \hbm_l &=  
          \softmax\left(\frac{\Xbf^Q W_l^Q  {W_l^K}^\top {\Xbf^K}^\top}{\sqrt{\dv}}\right) \Xbf^V W_l^V  \nonumber \\      
        &= \left[ \hbm_{l,1},\dots, \hbm_{l,N} \right]^\top \in \RR^{N \times \dv}, \\
    \hbm_{l,i} &=  
    \sum_{j = 1}^N  
    \frac{\exp\left(\frac{\xbm_i^\top W_l^Q  {W_l^K}^\top \xbm_j}{\sqrt{\dv}}\right)}{\sum_{k = 1}^N \exp\left(\frac{\xbm_i^\top W_l^Q  {W_l^K}^\top \xbm_k}{\sqrt{\dv}}\right)} {W_l^V}^\top \xbm_j.
\end{align}
We define the expert $f_j: \RR^{Nd} \rightarrow \RR^{d_v}$, and the score function $s_{i, j}: \RR^{Nd} \rightarrow \RR$ as follows:
\begin{align} 
    f_j(\boldsymbol{X}) &= W^{V^\top}_l E_j \boldsymbol{X} = W^{V^\top}_l\boldsymbol{x}_j, \label{eq:pretrain_expert} \\ 
    s_{i,j}(\boldsymbol{X}) &= \frac{\boldsymbol{X}^\top E_i^\top W_l^Q W_l^{K^\top}E_j \boldsymbol{X}}{\sqrt{d_v}} = \frac{\xbm_i^\top W_l^Q W_l^{K^\top}\xbm_j}{\sqrt{d_v}},
\end{align} 
for $i, j = 1, \dots, N$. Each output vector $\hbm_{l, i} \in \RR^{d_v}$ can then be expressed as an MoE model:
\begin{align}
    \hbm_{l,i} &= \sum^N_{j=1} \frac{\exp(s_{i,j}(\Xbm))}{\sum^N_{k=1} \exp({s_{i,k}}(\Xbm))}\cdot f_j(\Xbm).
\end{align}
This completes the proof of the proposition.

\subsection{Proof of Proposition~\ref{props: prompt_moe}} \label{appendix:proof_of_prompt_moe}

Similar to the notation in~\ref{appendix:proof_of_attn_moe}, we define the new experts and score functions for $j' = 1, \dots, L$ and $i = 1, \dots, N$ as follows:
\begin{align}
    f_{N+j'}(\boldsymbol{X}) &= W_l^{V^\top}\pbm_{j'}^V, \label{eq:prompt_expert}\\
    s_{i, N+j'}(\Xbm) &= \frac{\boldsymbol{X}^\top E_i^\top W_l^Q W_l^{K^\top} \pbm_{j'}^K}{\sqrt{d_v}} 
    = \frac{\xbm_i^\top W_l^Q W_l^{K^\top} \pbm_{j'}^K}{\sqrt{d_v}}.
\end{align}
Then, using Equation~\eqref{eq:prefix_tunining_output}, the output of the $l$-th attention head can be expressed as follows:
\begin{align}
    \htil_l &= \mathrm{Attention}\left(
    \Xbf^Q W_l^Q, 
    \begin{bmatrix}
            \Pbf^K \\ \Xbf^K
    \end{bmatrix} W_l^K, 
    \begin{bmatrix}
            \Pbf^V \\ \Xbf^V
    \end{bmatrix} W_l^V
    \right) = \left[ \htil_{l, 1}, \dots, \htil_{l, N} \right]^\top
    \in \RR^{N \times d_v}, \nonumber \\
    \htil_{l,i} 
    &= \sum_{j = 1}^N  
    \frac{\exp\left(\frac{\xbm_i^\top W_l^Q  {W_l^K}^\top \xbm_j}{\sqrt{\dv}}\right)}{\mathcal{D}_i(\Xbm)} {W_l^V}^\top \xbm_j 
    + \sum_{j' = 1}^L 
    \frac{\exp\left(\frac{\xbm_i^\top W_l^Q  {W_l^K}^\top \pbm_{j'}^K}{\sqrt{\dv}}\right)}
        {
            \mathcal{D}_i(\Xbm)
        } W_l^{V^\top}\pbm_{j'}^V
    \nonumber \\
    &= \sum_{j = 1}^N  
        \frac{\exp(s_{i, j}(\Xbm))}
        {
            \sum_{k = 1}^{N + L} \exp(s_{i, k}(\Xbm))
        } f_j(\Xbm)
    + \sum_{j' = 1}^L  
        \frac{\exp(s_{i, N + j'}(\Xbm))}
        {
            \sum_{k = 1}^{N + L} \exp(s_{i, k}(\Xbm))
        } f_{N + j'}(\Xbm),
\end{align}
where $\mathcal{D}_i(\Xbm) = 
\sum_{j = 1}^N 
    \exp\left(\frac{\xbm_i^\top W_l^Q  {W_l^K}^\top \xbm_j}{\sqrt{\dv}}\right)
+ \sum_{j' = 1}^L
    \exp\left(\frac{\xbm_i^\top W_l^Q  {W_l^K}^\top \pbm_{j'}^K}{\sqrt{\dv}}\right)$. This completes the proof of the proposition.

\section{Comparison to Sparse Mixture of Experts} \label{appendix:compare_smoe}

In this section, we detail how our task-specific prompt pool framework can be viewed as an implementation of a Sparse Mixture of Experts (SMoE) architecture. 

By setting the prompt length  $L = 1$, each prompt in the pool $\mathcal{P}_t$ (from Equation~\eqref{eq:prompt_pool}) functions as a single prefix expert. We denote these newly introduced experts as $f_{N + 1}^{(t)},\dots,f_{N + M}^{(t)}$. While the original $N$ experts from the pre-trained model $f_1,\dots,f_N$ are always active, we apply sparse selection exclusively to these new prefix experts.

Recall from Equation~\eqref{eq:prefix_moe} that each head in the MSA layer contains $N$ MoE models $\htil_{l, 1}, \dots, \htil_{l, N}$. A naive selection strategy would apply the $\mathrm{TopK}$ gating function independently to each of these $N$ models. This would require computing $N \times M$ score functions $s_{i, N + j'}(\Xbm)$ for all $i = 1, \dots, N$, $j' = 1, \dots, M$ and could result in a different set of prefix experts being selected for each model.

In contrast, our approach efficiently selects a single, shared set of $K$ new experts across all $N$ MoE models. We achieve this by using an auxiliary score function:
\begin{align}
    \Tilde{s}_{i, N + j'}(\Xbm) = \gamma(q(\xbm), \boldsymbol{k}_{j'}^{(t)}),
\end{align}
where $i=1,\dots,N$ and $j'=1,\dots,M$. Since $\Tilde{s}_{i, N + j'}(\Xbm)$ is independent of the index $i$ and depends only on the prompt key $\boldsymbol{k}_{j'}^{(t)}$, we only need to compute $M$ unique scores. This design significantly reduces the computational burden while enabling the efficient selection of top-$K$ experts from the prompt pool.

While computing the query feature $q(\xbm)$ might appear to be an additional expense, this feature is already required by our task predictor (Section~\ref{sec:method_gen_model}) during both training and inference. Therefore, we reuse this computation for prompt selection at no additional overhead.

\section{Comparison of Cascade Voting and the MLP-based Approach}\label{appendix:cascade_mlp}

Both the MLP-based head strategy proposed by \citet{le2024adaptive} and the cascade voting mechanism introduced by \citet{zhou2024ensemble} share the principle of treating each observed relation independently, rather than grouping all relations within a task into a single category. In the MLP-based paradigm, each head corresponds to a specific relation. In contrast, the cascade voting framework scores every relation within a task and selects the one with the highest score as that task's representative. Subsequently, the representatives from all tasks are assessed, and the task associated with the highest-scoring representative is designated as the predicted task.

A key distinction lies in \emph{whether task classification is explicitly trained}. The MLP-based paradigm employs a standard supervised approach: an MLP head is allocated to each relation, and the final prediction corresponds to the head with the highest output probability. In contrast, cascade voting dispenses with a trainable classifier, instead leveraging inference-time computations (\eg distance-based metrics) to identify the most appropriate task.

Regarding memory usage, the MLP-based approach only requires storing relation-specific distributions to facilitate data reconstruction and mitigate catastrophic forgetting. The cascade voting mechanism, however, relies on relation distributions that are further specialized by each prompt pool, resulting in a larger number of stored distributions. Despite this increased memory demand, a notable advantage of cascade voting is that it \emph{eliminates the overhead of training a task classifier}, thereby expediting the overall training process compared to its MLP-based counterpart.

\begin{table}[!t]
    \centering
    \begin{tabular}{l|c c c c c c c c c c}
    \hline
    \multicolumn{11}{c}{\textbf{FewRel}} \\
    \hline
    Num of descriptions  & $\mathcal{T}_1$ & $\mathcal{T}_2$ & $\mathcal{T}_3$ & $\mathcal{T}_4$ & $\mathcal{T}_5$ & $\mathcal{T}_6$ & $\mathcal{T}_7$ & $\mathcal{T}_8$ & $\mathcal{T}_9$ & $\mathcal{T}_{10}$ \\
    \hline 
    w/o descriptions &	96.9 & 94.0 & 92.6 & 91.7 & 90.9 & 89.8 & 88.7 & 88.2 & 87.2 & 85.8 \\
    
    1 description &	\textbf{98.2} &	\textbf{95.8} &	\textbf{95.1}   &	\textbf{94.1} & \textbf{92.7}   &	\textbf{91.9}   &	\underline{90.2} &	\textbf{89.9} &	\textbf{89.0} &	\textbf{87.7} \\
    3 descriptions &	97.7 & \underline{95.3} & \underline{94.1} & \underline{93.2} & 92.3 & \underline{91.4} & \underline{90.2} & 89.6 & 88.8 & 87.4 \\
    5 descriptions & 97.7 & \underline{95.3} & 94.0 & {93.1} & 92.3 & 91.4 & \textbf{90.3} & 89.7 & {88.8} & 87.5 \\
    7 descriptions &	\underline{97.8} & \underline{95.3} & \underline{94.1} & 93.1 & \underline{92.3} & {91.3} & \textbf{90.3} & \underline{89.8} & \underline{88.9} & \underline{87.6} \\
    \hline
    \hline
    \multicolumn{11}{c}{\textbf{TACRED}} \\
    \hline
    Num of descriptions  & $\mathcal{T}_1$ & $\mathcal{T}_2$ & $\mathcal{T}_3$ & $\mathcal{T}_4$ & $\mathcal{T}_5$ & $\mathcal{T}_6$ & $\mathcal{T}_7$ & $\mathcal{T}_8$ & $\mathcal{T}_9$ & $\mathcal{T}_{10}$ \\
    \hline 
    w/o descriptions &	97.4 &	93.0 &	89.1 &	86.0 &	84.4 &	83.5 &	82.8 &	81.8 &	81.1 &	80.7 \\
    
    1 description &	\underline{97.6} &	{93.6} &	\underline{90.7} &	{88.2} & \underline{86.4} & \textbf{85.4} & \textbf{84.3} &\textbf{83.7} & \textbf{83.2} &	\textbf{82.5} \\
    3 descriptions &	97.5 &	\textbf{94.0} & \textbf{90.8} &	\textbf{88.8}  &	\underline{86.4} &	{84.8} &	84.1 &	82.9 &	\underline{82.6} &	81.9 \\
    5 descriptions &	97.3 &	\underline{93.9} &	\underline{90.7} &	87.7 &	86.3 &	84.4 &	83.9 &	\underline{83.2} &	82.6 &	\underline{82.1} \\
    7 descriptions &	\textbf{97.7} &	93.4 &	90.3 &	\underline{88.5} &	\textbf{86.8} &	\underline{85.2} &	\underline{84.2} &	83.1 &	82.5 &	82.1 \\
    \hline
    \end{tabular}
    \caption{Detailed analyses of WAVE++ with different number of descriptions. For each dataset and metric, the best result is \textbf{bolded} and the second is \underline{underline}.}
    \label{tab:num_des}
    \vskip -0.1in
\end{table}

\section{Impact of the Number of Label Descriptions $D$} \label{appendix:num_desc}

To evaluate the sensitivity of WAVE++ to the number of label descriptions $D$, as defined in Equation~\eqref{eq:des_loss}, we conducted experiments on the FewRel and TACRED datasets, varying $D$ from 0 to 7. The results, presented in Table~\ref{tab:num_des}, demonstrate that WAVE++ maintains robust performance even with a limited number of descriptions.

These findings indicate that WAVE++ is not heavily dependent on a large set of label descriptions to perform effectively. Indeed, the model achieves high accuracy with only a few descriptions, offering a favorable trade-off between performance and computational efficiency. This robustness makes WAVE++ highly suitable for practical applications where acquiring extensive label descriptions is either infeasible or costly.

\section{Gaussian Justification}

\begin{figure}[ht]
    \centering
    \includegraphics[width=0.75\linewidth]{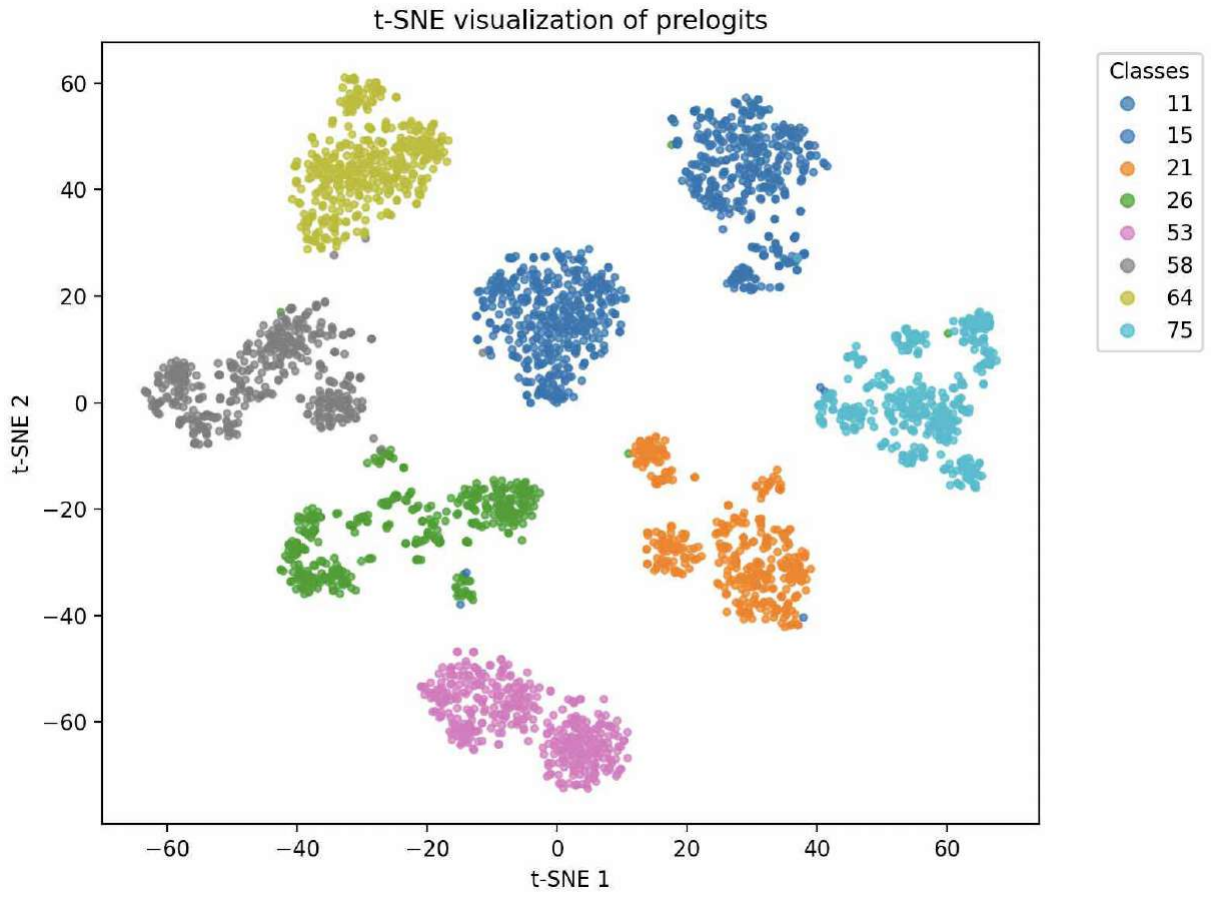}
    \caption{t-SNE visualization of hidden representations from a representative task in the FewRel dataset.}
    \label{fig:gauss}
\end{figure}

In WAVE++, we approximate the distribution of prompted representations within each task using a Gaussian model. To empirically validate this approximation, we present a t-SNE visualization of the prompted representations $f_\theta(\xbm_p)$ for a representative task from the FewRel dataset, as shown in Figure~\ref{fig:gauss}. The visualization reveals clear clustering patterns in the latent space, suggesting that a Gaussian distribution provides a reasonable approximation of the underlying structure.

{To more rigorously assess the validity of the Gaussian assumption, we conduct Royston’s test \citep{royston1992approximating} for multivariate normality. Royston’s test extends the Shapiro–Wilk test to the multivariate setting by aggregating dimension-wise Shapiro–Wilk statistics and comparing the resulting test statistic to a chi-square distribution. The null hypothesis $H_0$ assumes that the data are drawn from a multivariate Gaussian distribution, while the alternative hypothesis $H_1$ assumes otherwise. We adopt a significance level of $\alpha = 0.05$.

\begin{table}[htbp]
    \centering
    \begin{tabular}{l |c c c c c c c c}
    \hline 
    Labels & 11 &15 &21 & 26 & 53& 58& 64& 75 \\
    \hline
    $p$-value & 0.87 & 0.68 & 0.39 & 0.09 & 0.57 & 0.42 & 0.65 & 0.38\\
    \hline
    \end{tabular}
    \caption{{Royston's test on samples from eight classes within a task in FewRel, demonstrating that data can be approximated as Gaussian distribution.}}
    \label{tab: gauss-test}
\vspace{-1.0em}
\end{table}

The test is performed on a representative FewRel task containing eight relation classes. As reported in Table~\ref{tab: gauss-test}, all classes yield $p$-values greater than $0.05$, indicating no statistically significant evidence against the multivariate normality assumption. These results provide empirical support for the use of Gaussian modeling in our latent space.

More broadly, our objective is to map raw inputs into a representation space where class separation can be effectively achieved using linear decision boundaries. Within this paradigm, Gaussian modeling is a practical and widely adopted choice, as demonstrated in prior continual learning methods such as HiDe-Prompt and EoE. Moreover, we employ a task-specific shared covariance matrix $\boldsymbol{\Sigma}^t$ across all relations, which substantially reduces the number of parameters and improves scalability without sacrificing empirical performance.}

\section{Performance under Incorrect Task Prediction}

\begin{table}[!t]
\centering
\resizebox{\textwidth}{!}{
\begin{tabular}{llcccccccccc}
\toprule
\multicolumn{12}{c}{\textbf{FewRel}}                                                                          \\ \midrule
\multicolumn{1}{l|}{Model}                      & \multicolumn{1}{l|}{}                                      & $\mathcal{T}_1$                          & $\mathcal{T}_2$                             & $\mathcal{T}_3$                             & $\mathcal{T}_4$                             & $\mathcal{T}_5$                             & $\mathcal{T}_6$                             & $\mathcal{T}_7$                             & $\mathcal{T}_8$                             & $\mathcal{T}_9$                             & $\mathcal{T}_{10}$                          \\  \hline
\multicolumn{1}{c|}{}                           & \multicolumn{1}{l|}{Percentage}                            & 0                                 & {\color[HTML]{3C4043} 18.2}          & {\color[HTML]{3C4043} 10.0}          & {\color[HTML]{3C4043} 13.8}          & {\color[HTML]{3C4043} 11.8}          & {\color[HTML]{3C4043} 9.4}           & {\color[HTML]{3C4043} 11.8}          & {\color[HTML]{3C4043} 11.9}          & {\color[HTML]{3C4043} 12.4}          & {\color[HTML]{3C4043} 14.4}          \\
\multicolumn{1}{c|}{}                           & \multicolumn{1}{l|}{Correct final prediction cases} & {\color[HTML]{3C4043} 0}          & {\color[HTML]{3C4043} 4}             & {\color[HTML]{3C4043} 4}             & {\color[HTML]{3C4043} 8}             & {\color[HTML]{3C4043} 8}             & {\color[HTML]{3C4043} 8}             & {\color[HTML]{3C4043} 13}            & {\color[HTML]{3C4043} 14}            & {\color[HTML]{3C4043} 16}            & {\color[HTML]{3C4043} 20}            \\
\multicolumn{1}{c|}{\multirow{-3}{*}{WAVE-CRE}} & \multicolumn{1}{l|}{Errors in task prediction}   & {\color[HTML]{3C4043} 0}          & {\color[HTML]{3C4043} 22}            & {\color[HTML]{3C4043} 40}            & {\color[HTML]{3C4043} 58}            & {\color[HTML]{3C4043} 68}            & {\color[HTML]{3C4043} 85}            & {\color[HTML]{3C4043} 110}           & {\color[HTML]{3C4043} 118}           & {\color[HTML]{3C4043} 129}           & {\color[HTML]{3C4043} 139}           \\ \hline
\multicolumn{1}{l|}{}                           & \multicolumn{1}{l|}{Percentage}                            & {\color[HTML]{3C4043} \textbf{0}} & {\color[HTML]{3C4043} \textbf{39.3}} & {\color[HTML]{3C4043} \textbf{22.0}} & {\color[HTML]{3C4043} \textbf{17.1}} & {\color[HTML]{3C4043} \textbf{18.2}} & {\color[HTML]{3C4043} \textbf{17.3}} & {\color[HTML]{3C4043} \textbf{18.4}} & {\color[HTML]{3C4043} \textbf{20.5}} & {\color[HTML]{3C4043} \textbf{15.3}} & {\color[HTML]{3C4043} \textbf{19.2}} \\
\multicolumn{1}{l|}{}                           & \multicolumn{1}{l|}{Correct final prediction cases} & {\color[HTML]{3C4043} 0}          & {\color[HTML]{3C4043} 7}             & {\color[HTML]{3C4043} 8}             & {\color[HTML]{3C4043} 8}             & {\color[HTML]{3C4043} 11}            & {\color[HTML]{3C4043} 15}            & {\color[HTML]{3C4043} 18}            & {\color[HTML]{3C4043} 22}            & {\color[HTML]{3C4043} 18}            & {\color[HTML]{3C4043} 25}            \\
\multicolumn{1}{l|}{\multirow{-3}{*}{WAVE++}}   & \multicolumn{1}{l|}{Errors in task prediction}   & {\color[HTML]{3C4043} 0}          & {\color[HTML]{3C4043} 18}            & {\color[HTML]{3C4043} 37}            & {\color[HTML]{3C4043} 47}            & {\color[HTML]{3C4043} 60}            & {\color[HTML]{3C4043} 89}            & {\color[HTML]{3C4043} 102}           & {\color[HTML]{3C4043} 110}           & {\color[HTML]{3C4043} 119}           & {\color[HTML]{3C4043} 133}           \\ \hline
\hline
\multicolumn{12}{c}{\textbf{TACRED}}                                                                                                                                                                                                                                                                                                                                                                                                                                                                            \\ \hline
\multicolumn{1}{l|}{Model}                      & \multicolumn{1}{l|}{}                                      & $\mathcal{T}_1$                          & $\mathcal{T}_2$                             & $\mathcal{T}_3$                             & $\mathcal{T}_4$                             & $\mathcal{T}_5$                             & $\mathcal{T}_6$                             & $\mathcal{T}_7$                             & $\mathcal{T}_8$                             & $\mathcal{T}_9$                             & $\mathcal{T}_{10}$                          \\ \hline
\multicolumn{1}{l|}{}                           & \multicolumn{1}{l|}{Percentage}                            & {\color[HTML]{3C4043} 0}          & {\color[HTML]{3C4043} 14.3}          & {\color[HTML]{3C4043} 5.0}           & {\color[HTML]{3C4043} 15.0}          & {\color[HTML]{3C4043} 10.7}          & {\color[HTML]{3C4043} 7.7}           & {\color[HTML]{3C4043} 8.0}           & {\color[HTML]{3C4043} 7.7}           & {\color[HTML]{3C4043} 8.3}           & {\color[HTML]{3C4043} 12.0}          \\
\multicolumn{1}{l|}{}                           & \multicolumn{1}{l|}{Correct final prediction cases} & {\color[HTML]{3C4043} 0}          & {\color[HTML]{3C4043} 1}             & {\color[HTML]{3C4043} 1}             & {\color[HTML]{3C4043} 3}             & {\color[HTML]{3C4043} 3}             & {\color[HTML]{3C4043} 2}             & {\color[HTML]{3C4043} 2}             & {\color[HTML]{3C4043} 2}             & {\color[HTML]{3C4043} 2}             & {\color[HTML]{3C4043} 3}             \\
\multicolumn{1}{l|}{\multirow{-3}{*}{WAVE-CRE}} & \multicolumn{1}{l|}{Errors in task prediction}   & {\color[HTML]{3C4043} 0}          & {\color[HTML]{3C4043} 7}             & {\color[HTML]{3C4043} 20}            & {\color[HTML]{3C4043} 20}            & {\color[HTML]{3C4043} 28}            & {\color[HTML]{3C4043} 26}            & {\color[HTML]{3C4043} 25}            & {\color[HTML]{3C4043} 26}            & {\color[HTML]{3C4043} 24}            & {\color[HTML]{3C4043} 25}            \\ \hline
\multicolumn{1}{l|}{}                           & \multicolumn{1}{l|}{Percentage}                            & {\color[HTML]{3C4043} 0}          & {\color[HTML]{3C4043} \textbf{25.0}} & {\color[HTML]{3C4043} \textbf{25.0}} & {\color[HTML]{3C4043} \textbf{26.3}} & {\color[HTML]{3C4043} \textbf{19.0}} & {\color[HTML]{3C4043} \textbf{14.3}} & {\color[HTML]{3C4043} \textbf{16.7}} & {\color[HTML]{3C4043} \textbf{16.7}} & {\color[HTML]{3C4043} \textbf{23.5}} & {\color[HTML]{3C4043} \textbf{20.0}} \\
\multicolumn{1}{l|}{}                           & \multicolumn{1}{l|}{Correct final prediction cases} & {\color[HTML]{3C4043} 0}          & {\color[HTML]{3C4043} 2}             & {\color[HTML]{3C4043} 5}             & {\color[HTML]{3C4043} 5}             & {\color[HTML]{3C4043} 4}             & {\color[HTML]{3C4043} 3}             & {\color[HTML]{3C4043} 3}             & {\color[HTML]{3C4043} 3}             & {\color[HTML]{3C4043} 4}             & {\color[HTML]{3C4043} 3}             \\
\multicolumn{1}{l|}{\multirow{-3}{*}{WAVE++}}   & \multicolumn{1}{l|}{Errors in task prediction}   & {\color[HTML]{3C4043} 0}          & {\color[HTML]{3C4043} 8}             & {\color[HTML]{3C4043} 20}            & {\color[HTML]{3C4043} 19}            & {\color[HTML]{3C4043} 21}            & {\color[HTML]{3C4043} 21}            & {\color[HTML]{3C4043} 18}            & {\color[HTML]{3C4043} 18}            & {\color[HTML]{3C4043} 17}            & {\color[HTML]{3C4043} 15}            \\ \hline
\end{tabular}}
\caption{Final accuracy under incorrect task prediction}
\label{table:taskerror}
\vskip -0.1in
\end{table}

We present detailed results on the model's performance when the initial task prediction is incorrect. Table~\ref{table:taskerror} reports three key metrics: (i) the number of task prediction errors, (ii) the number of correct final relation predictions despite an incorrect task prediction, and (iii) the percentage of correct predictions among all cases of incorrect task prediction.

Across both datasets, WAVE++ consistently yields fewer task prediction errors than WAVE-CRE, highlighting the advantage of cascade voting over the MLP classifier. Moreover, when the task prediction is incorrect, WAVE++ achieves substantially higher final classification accuracy, outperforming WAVE-CRE by 3–10\% on FewRel and 7–20\% on TACRED. These results demonstrate the robustness of WAVE++ with cascade voting, even in the presence of task prediction errors.

\section{Statistical Significance Tests}

{To assess whether the proposed method WAVE++ significantly outperforms prior approaches, we conducted paired t-tests against two baselines, WAVE-CRE and HiDe-Prompt, on the TACRED and FewRel datasets. Each comparison was performed over five independent runs with different random seeds.}

{
For each test, the null hypothesis $H_0$ states that the mean performance difference between WAVE++ and the corresponding baseline is zero, while the alternative hypothesis $H_1$ posits that WAVE++ outperforms the baseline. We adopt a significance level of $\alpha = 0.05$. As reported in Table~\ref{tab: t-test}, all obtained p-values are substantially below the significance threshold, demonstrating that the performance gains of WAVE++ over both WAVE-CRE and HiDe-Prompt are \emph{statistically significant}.}

\begin{table}[htbp]
    \centering
    {\begin{tabular}{l | c c c | c c c}
    \hline 
    Dataset & \multicolumn{3}{c|}{\textbf{TACRED}} & \multicolumn{3}{c}{\textbf{FewRel}} \\
    \hline
    Method  & WAVE-CRE & HiDe-Prompt & EoE& WAVE-CRE & HiDe-Prompt &EoE\\
    \hline
    $p$-value & 3.6e-04 & 9.5e-04 & 5.4e-03 & 2.8e-06 & 5.2e-04 & 1.5e-04\\
    \hline

    \end{tabular}}
    \caption{Paired t-test results evaluating whether WAVE++ significantly outperforms on the TACRED and FewRel datasets.}
    \label{tab: t-test}
\end{table}

\section{Time Analysis} \label{appendix:time_analysis}

\begin{table}[ht]
\centering
{\begin{tabular}{@{}l|cc|cc@{}}
\toprule
\multirow{2}{*}{Method} & \multicolumn{2}{c|}{Training Time (h)} & \multicolumn{2}{c}{Inference Latency (ms)} \\ \cmidrule(l){2-5} 
                        & TACRED             & FewRel            & TACRED               & FewRel              \\ \midrule
L2P                     & 0.9                & 6.1               & 26.6                 & 27.2                \\
EoE                     & 1.2                & 6.5              & 39.5                 & 40.6 \\
WAVE-CRE                & 2.5                & 7.4              &         28.7                 & 29.8                       \\
WAVE++                  & 1.4                & 6.8               & 40.5                 & 41.2                \\ \bottomrule
\end{tabular}}
    \caption{Training time and inference latency on the TACRED and FewRel datasets.}
    \label{tab: time}
\end{table}

{In this section, we evaluate the computational efficiency of the proposed method by comparing training time and inference latency, as reported in Table~\ref{tab: time}. As shown, WAVE++ achieves training time and inference latency comparable to the state-of-the-art method EoE. Moreover, both EoE and WAVE++ require substantially less training time than WAVE-CRE, as they eliminate the need to train a task predictor during the learning phase. This efficiency gain, however, comes at the cost of increased inference latency, since cascade voting is employed at test time. This trade-off between training efficiency and inference overhead is shared by both EoE and WAVE++.}


\end{document}